\documentclass[runningheads]{llncs}

\usepackage{eccv}

\usepackage{eccvabbrv}

\usepackage{graphicx}
\usepackage{booktabs}

\usepackage[accsupp]{axessibility}  % Improves PDF readability for those with disabilities.

\usepackage{float}
 \usepackage{algorithmic}
 \usepackage{graphicx}
 \usepackage{textcomp}
 \usepackage{xcolor}
 \usepackage{hyperref}
 \usepackage{adjustbox}
 \usepackage{amsmath}  % For mathematical environments
 \usepackage{tikz} % \usepackage{inconsolata}
  \usepackage{booktabs}
   \usepackage{multirow}
   \usepackage{graphicx}
 \usepackage{algorithm}
 \usepackage[normalem]{ulem}

 \usepackage{pifont}
   \usepackage{amssymb}   % for \checkmark
 \usepackage{makecell}  % for multi-line cells

 \newcommand{\xmark}{\ding{55}} % Define \xmark as a cross

 \newcommand{\cmark}{\checkmark}%

\newcommand{\maketitlesupplementary}{%
    \begin{center}
        {\LARGE\bfseries Supplementary Material\par}
    \end{center}
    \vspace{1em}
}

\usepackage{hyperref}

\usepackage{orcidlink}

\begin{document}

\title{GIF: A Conditional Multimodal Generative Framework for IR Drop Imaging in Chip Layouts} 

\titlerunning{Abbreviated paper title}

% \author{Kiran Thorat \inst{1}\orcidlink{0000-1111-2222-3333} \and
% Second Author\inst{2,3}\orcidlink{1111-2222-3333-4444} \and
% Third Author\inst{3}\orcidlink{2222--3333-4444-5555}}

% % TODO FINAL: Replace with an abbreviated list of authors.

% % TODO FINAL: Replace with your institution list.
% Springer Heidelberg, Tiergartenstr.~17, 69121 Heidelberg, Germany

\author{
Kiran Thorat\inst{1} \and
Nicole Meng\inst{2} \and
Mostafa Karami\inst{1} \and
Caiwen Ding\inst{3} \and
Yingjie Lao\inst{2} \and
Zhijie Jerry Shi\inst{1}
}

\authorrunning{K. Thorat et al.}

\institute{
University of Connecticut, USA\\
\email{\{kiran\_gautam.thorat, mostafa.karami, zhijie.shi\}@uconn.edu}
\and
Tufts University, USA\\
\email{\{ziyi.meng, Yingjie.Lao\}@tufts.edu}
\and
University of Minnesota Twin Cities, USA\\
\email{dingc@umn.edu}
}

\maketitle

\begin{abstract}
IR drop analysis is essential in physical chip design to ensure the power integrity of on-chip power delivery networks. Traditional Electronic Design Automation (EDA) tools have become slow and expensive as transistor density scales. Recent works have introduced machine learning (ML)-based methods that formulate IR drop analysis as an image prediction problem. These existing ML approaches fail to capture both local and long-range dependencies and ignore crucial geometrical and topological information from physical layouts and logical connectivity. To address these limitations, we propose GIF, a \textbf{\textit{\underline{G}}}enerative \textbf{\textit{\underline{I}}}R drop \textbf{\textit{\underline{F}}}ramework that uses both geometrical and topological information to generate IR drop images. GIF fuses image and graph features to guide a conditional diffusion process, producing high-quality IR drop images. For instance, 
On the CircuitNet-N28 dataset, GIF achieves 0.78 SSIM, 0.95 Pearson correlation, 21.77 PSNR, and 0.026 NMAE, outperforming prior methods.
These results demonstrate that our framework, using diffusion based multimodal conditioning, reliably generates high quality IR drop images. This shows that IR drop analysis can effectively leverage recent advances in generative modeling when geometric layout features and logical circuit topology are jointly modeled. By combining geometry aware spatial features with logical graph representations, GIF enables IR drop analysis to benefit from recent advances in generative modeling for structured image generation.
\end{abstract}    
\section{Introduction}
\label{sec:intro}

As semiconductor technology nodes continue to shrink and on-chip transistor density increases, the resulting growth in layout complexity makes IR-drop simulation increasingly computationally expensive~\cite{zhao2024pdnnetpdnawaregnncnnheterogeneous, jiang2024circuitnet,chai2023circuitnet}. ML-based IR-drop prediction methods have therefore been explored as a promising solution. Most existing approaches rely on convolutional neural networks (CNNs) or graph neural networks (GNNs)~\cite{9045303,9045574,fang2018machine,ho2019incpird}. For example, PDNNet~\cite{zhao2024pdnnetpdnawaregnncnnheterogeneous} uses a GNN--CNN architecture in which the power delivery network (PDN) is modeled as a graph and localized spatial features (e.g., internal power) are extracted using a CNN. However, the limited receptive field of CNNs hampers their ability to capture the global spatial dependencies present in modern chip layouts~\cite{zheng2023lay}.
Existing ML based approaches typically model either spatial layout features or circuit connectivity, but rarely integrate both sources of information in a unified representation. 
 GNN--CNN hybrids such as PDNNet incorporate PDN structure and power features, but do not model richer topological information such as logical connectivity between cells and nets. Existing image-based approaches, including PowerNet~\cite{Xie_2020} and MAVIREC~\cite{chhabria2021mavirec}, treat IR-drop prediction as a spatial regression problem and focus primarily on power-related maps, while omitting additional geometric features (e.g., cell-density and placement-derived spatial indicators) that correlate with local current demand and local variations in IR-drop. As a result, existing methods struggle to jointly capture (i) fine-grained geometric variation across the layout and (ii) long-range dependencies induced by the logical structure of the design.

Our key insight is that IR-drop patterns are inherently \emph{multimodal}: they depend simultaneously on local geometric context (e.g., spatial distribution of power) and global topological dependencies (e.g., logical connectivity that influences correlated current demand). Rather than directly regressing IR-drop values from limited features, we view IR-drop prediction as a \emph{conditional generative} problem. By learning a generative model conditioned on both geometric features and a graph representation of the design, the model can capture the underlying distribution of IR-drop maps that are consistent with both the layout geometry and the logical structure.
% \begin{table}[t]
% \centering
% \caption{IR-drop prediction/generation comparison.}
% \scriptsize
% \begin{tabular}{lcccc}
% \toprule
% \textbf{Method} &
% \textbf{\begin{tabular}[c]{@{}c@{}}Power\\ Map\end{tabular}} &
% \textbf{\begin{tabular}[c]{@{}c@{}}Topology\\ Info\end{tabular}} &
% \textbf{\begin{tabular}[c]{@{}c@{}}Geometric\\ Features\end{tabular}} &
% \textbf{\begin{tabular}[c]{@{}c@{}}Generative\\ Model\end{tabular}} \\
% \midrule
% PowerNet~\cite{Xie_2020} & \checkmark & \xmark & \xmark & \xmark \\
% MAVIREC~\cite{chhabria2021mavirec} & \checkmark & \xmark & \xmark & \xmark \\
% PDNNet~\cite{zhao2024pdnnetpdnawaregnncnnheterogeneous} & \checkmark & \checkmark & \xmark & \xmark \\
% \midrule
% \textbf{Ours (GIF)} & \checkmark & \checkmark & \checkmark & \checkmark \\
% \bottomrule
% \end{tabular}
% \label{tab:comparison_methods}
% \end{table}
To address these limitations, we propose GIF, a diffusion-based generative framework that fuses geometric information and topological information for conditional IR-drop map generation. On the geometric side, we construct a multi-channel image representation that augments conventional power maps with additional spatial features that better characterize variability across the layout. On the topological side, we build a graph from the logical netlist, encode it using a GNN, and compress the resulting node embeddings into a fixed set of graph tokens that summarize global logical structure. A diffusion U-Net is then conditioned on both modalities: geometric features modulate the network via feature-wise conditioning. This allows GIF to effectively integrate local detail with long-range dependencies during IR-drop generation. Table~\ref{tab:comparison_methods} summarizes key differences between existing IR-drop prediction approaches and GIF.
\begin{table}[t]
\centering
\caption{Comparison of IR-drop learning approaches.}
\scriptsize
\setlength{\tabcolsep}{5pt}
\begin{tabular}{lcccc}
\toprule
\textbf{Method} &
\textbf{\begin{tabular}[c]{@{}c@{}}Layout\\ Power Maps\end{tabular}} &
\textbf{\begin{tabular}[c]{@{}c@{}}Geometry-aware\\ Layout Features\end{tabular}} &
\textbf{\begin{tabular}[c]{@{}c@{}}Logical\\ Topology\end{tabular}} &
\textbf{\begin{tabular}[c]{@{}c@{}}Generative\\ Modeling\end{tabular}} \\
\midrule
PowerNet~\cite{Xie_2020} & \checkmark & \xmark & \xmark & \xmark \\
MAVIREC~\cite{chhabria2021mavirec} & \checkmark & \xmark & \xmark & \xmark \\
PDNNet~\cite{zhao2024pdnnetpdnawaregnncnnheterogeneous} & \checkmark & \xmark & \checkmark & \xmark \\
\midrule
\textbf{Ours (GIF)} & \checkmark & \checkmark & \checkmark & \checkmark \\
\bottomrule
\end{tabular}
\label{tab:comparison_methods}
\end{table}
Our key contributions are summarized below:

\begin{enumerate}
\item \textbf{Multimodal generative formulation for IR-drop analysis.}
We formulate IR-drop prediction as a conditional generation of IR-drop maps that jointly models geometric layout structure and logical circuit connectivity. This formulation enables learning IR-drop maps consistent with both local spatial features and global circuit dependencies.
    \item \textbf{Geometry-enhanced spatial features.} Beyond conventional power maps, we incorporate additional spatial features derived from placement, enabling the model to account for geometric variations that influence IR-drop behavior. This produces a improved dataset with richer spatial representation with additional structural cues that improve IR-drop modeling. 
    % This produces an improved dataset with richer structural cues, supporting more reliable evaluation.
    \item \textbf{Topology-aware logical graph modeling.} We construct a netlist-level graph that encodes logical connectivity between cells and nets, and obtain a fixed set of graph tokens using a GNN encoder and token-pooling scheme to capture global structural dependencies.
\item \textbf{Multimodal fusion of physical-design features for IR-drop generation.}
We design a diffusion-based IR-drop generator in which geometric features provide spatial conditioning and logical graph tokens are injected via gated cross-attention, allowing the denoising network to incorporate circuit-level context throughout the generative process.
\end{enumerate}
Our proposed framework, GIF, achieves 0.786 SSIM, 0.9536 Pearson correlation, 21.77 PSNR, and 0.0266 NMAE, outperforming all existing methods across these metrics. PowerNet~\cite{Xie_2020} reports 0.56 SSIM, 0.77 correlation, 11.60 PSNR, and 0.149 NMAE; MAVIREC~\cite{chhabria2021mavirec} reports 0.68 SSIM, 0.91 correlation, 18.27 PSNR, and 0.039 NMAE; and PDNNet~\cite{zhao2024pdnnetpdnawaregnncnnheterogeneous} reports 0.72 SSIM, 0.92 correlation, 19.35 PSNR, and 0.028 NMAE. To the best of our knowledge, GIF is the first diffusion-based framework for IR-drop map generation that jointly conditions on geometric layout features and logical graph topology.

\section{Background and Related Work}

\noindent\textbf{CNN-based IR-drop models.}
CNN-based learning IR-drop prediction is typically formulated as an image-to-image regression problem. Additional background on chip design flow and IR-drop analysis is provided in the supplementary material~\ref{sec:supp_background}. 
% (Background: Modern Chip Design Flow and IR-Drop).
PowerNet~\cite{Xie_2020} employs a U-Net to map layout features to IR-drop images, MAVIREC~\cite{chhabria2021mavirec} improves hotspot localization, and PDNNet~\cite{zhao2024pdnnetpdnawaregnncnnheterogeneous} integrates CNN-GNN spatial modeling.  These models rely solely on image features and struggle to capture long-range dependencies in power-delivery networks.
\medskip

\noindent\textbf{GNN-based chip-design models.}
GNNs are widely applied in physical-design tasks to represent logical connectivity and structural relationships~\cite{yang2022versatile,thoratgroot}.  
Despite their relevance to global PDN behavior, existing IR-drop models do not incorporate topological information from the netlist.
While GNNs capture netlist structure effectively, most existing approaches apply them to node-level voltage estimation or global PDN analysis rather than generating full-resolution IR-drop maps.
% This leaves a disconnect between topology-aware modeling and pixel-level IR-drop behavior, since graph signals are rarely fused with layout-derived image features in a unified model.

% \medskip
% \noindent\textbf{Transformers and generative models.}
% Transformers capture global spatial interactions~\cite{dosovitskiy2020image,liu2021swin}, and layout-specific variants such as Lay-Net~\cite{zheng2023lay} demonstrate advantages for chip layout for congestion prediction.  
% Generative models including VAEs~\cite{kingma2013auto}, denoising diffusion models~\cite{ho2020denoising}, and latent diffusion~\cite{rombach2022high} have achieved best performance in image synthesis.  
% However, prior IR-drop work remains deterministic and does not exploit generative modeling or multimodal conditioning.

\medskip
\noindent\textbf{Transformers and generative models.}
Transformers capture global spatial interactions~\cite{dosovitskiy2020image,liu2021swin}, and layout-specific variants such as Lay-Net~\cite{zheng2023lay} demonstrate advantages for chip layout for congestion prediction.  
Generative models including VAEs~\cite{kingma2013auto}, denoising diffusion models~\cite{ho2020denoising}, and latent diffusion~\cite{rombach2022high} have achieved best performance in image synthesis.  
However, prior IR-drop work remains deterministic and does not exploit generative modeling or multimodal conditioning.  
This work introduces the first generative formulation for IR-drop, filling a gap in prior methods.
\subsection{Problem Formulation}
Given a placed digital design, let $\mathbf{X}_{p} \in \mathbb{R}^{H \times W \times C_{p}}$ denote the set of physical layout features, including power-map features and geometry-aware quantities such as cell density, RUDY, and overflow. Let $\mathcal{G} = (\mathcal{V}, \mathcal{E})$ denote the topology-aware graph constructed from the netlist, where each node $v \in \mathcal{V}$ represents a cell instance and edges $e \in \mathcal{E}$ represent logical connectivity.
We define the mapping
\begin{equation}
f_{\theta}: \left(\mathbf{X}_{p}, \mathcal{G}\right) \rightarrow \mathbf{Y},
\end{equation}
where $\mathbf{Y} \in \mathbb{R}^{H \times W}$ is the IR-drop map.
In the generative formulation, we instead model the conditional distribution
\begin{equation}
p_{\theta}\left(\mathbf{Y} \mid \mathbf{X}_{p}, \mathcal{G}\right),
\end{equation}
and generate samples of $\mathbf{Y}$ by drawing from this distribution.
The problem is therefore to estimate the conditional distribution $p_{\theta}\left(\mathbf{Y} \mid \mathbf{X}_{p}, \mathcal{G}\right)$ and generate IR-drop maps that match the ground-truth distribution.
\begin{figure*}[t]
    \centering
    \includegraphics[width=0.98\linewidth]{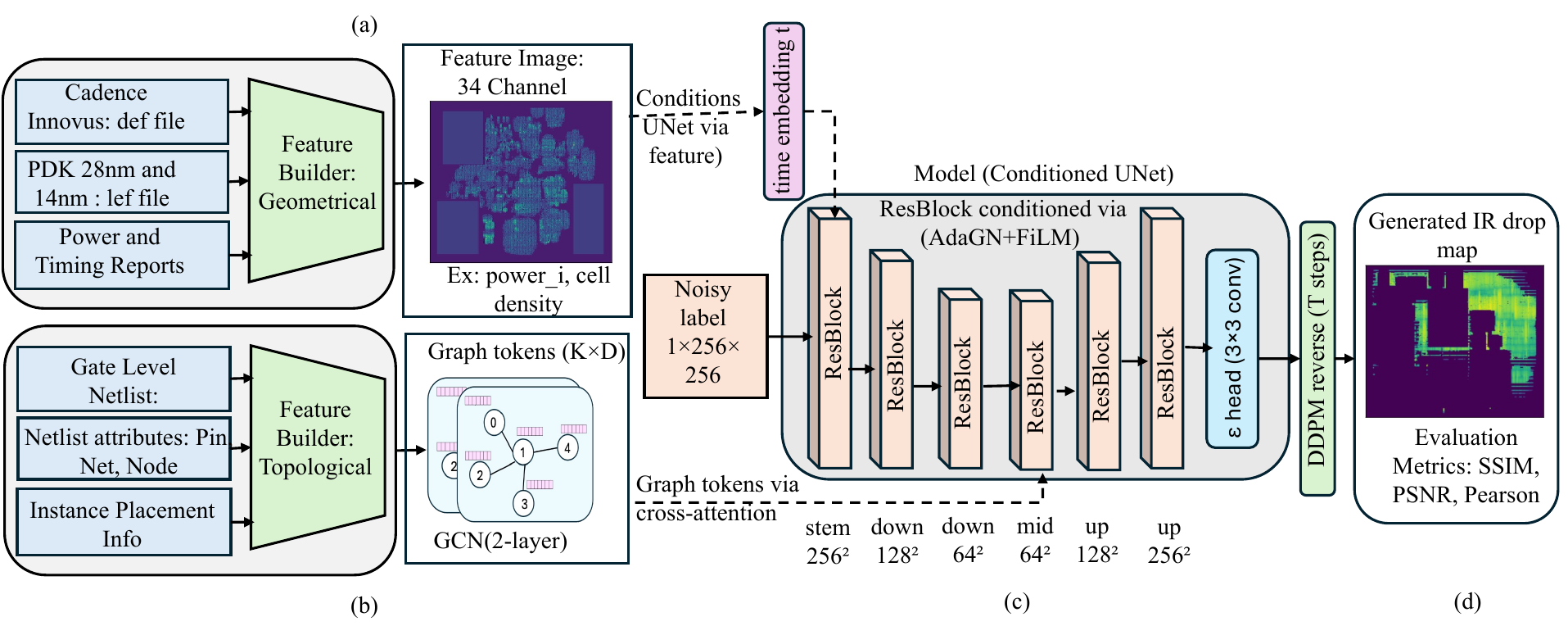}
    \caption{
        Overview of the proposed framework. (a) Geometric features creation from DEF/LEF files and power reports, (b) Topological features creation, and
        (c) A diffusion-based UNet predicts the noise $\epsilon$, conditioned on features via AdaGN+FiLM and on graph tokens via gated cross-attention, (d) Generated IR drop map.
    }
    \label{fig:overview}
\end{figure*}
\section{Framework}
\label{sec:frame}

% \hl{suggest to put together an algorithm for the entire method}

Figure ~\ref{fig:overview} provides an overview of the proposed framework GIF. On the left side of the figure, two feature-builder modules generate the conditioning inputs. The upper block (Figure ~\ref{fig:overview} (a)) processes def/lef files, PDK information, and power/timing reports to form a multi-channel geometry-aware feature image (maps). The lower block (Figure ~\ref{fig:overview} (b)) uses the gate-level netlist together with pin, net, node attributes and instance information to form topology (logical connection) aware feature. As shown in the figure, this block applies a lightweight two-layer GCN followed by a top-$k$ selection step, producing a fixed-size set of graph tokens that represent logical connectivity.
The diffusion U-Net appears on the right half of Figure~\ref{fig:overview} (c). The noisy IR-drop label enters the first ResBlock, and the sequence of ResBlocks corresponds directly to the stack shown in dark (UNet) block. Geometry-aware maps condition every ResBlock through AdaGN and FiLM layers, as indicated by the conditioning arrows. Graph tokens are injected through gated cross-attention blocks placed at the bottleneck and early decoder positions, as marked in Figure~\ref{fig:overview}. A sinusoidal time embedding is added to each block. The final 3×3 convolutional head produces the noise estimate, and the DDPM reverse process generates the IR-drop map shown in Figure~\ref{fig:overview} (d). The next subsections describe the geometry-aware feature maps, the topology-aware graph construction, and the multimodal conditioning mechanism in details.

\subsection{Construction of geometry-aware feature maps}
IR drop analysis is a simulation of voltage drop under the combined effect of parasitic and current  flow through the on-chip power delivery network (PDN) \cite{zhong2005fast}. It involves large system of equations of the form \cite{kose2011fast},
\begin{equation}
G\,v = i,
\label{eq:gv_i}
\end{equation}
where $G$ is the transconductance matrix,  $v$ is the voltage at the pin, and $i$ is the current flowing through the PDN. 
The modeling of IR drop using the system of equations is expensive and slow. To address these challenges most ML-based IR drop prediction methods use the tile based power maps (feature images) \cite{chhabria2021mavirec}. Based on these IR drop feature images, ML based modes predict the IR drop.   
These feature images are created based on the power reports and timing window reports.
Power reports contains instance level Internal power ($p_i$), Switching power ($p_s$), Leakage power ($p_l$), Toggle rate ($r_{togg}$). 

For each layout tile, instance-level internal power ($p_i$), switching power ($p_s$), leakage power ($p_l$), and toggle rate ($r_{\text{togg}}$) are aggregated to form several normalized channels:
\begin{align}
p_{\text{i}} &\propto p_i, \qquad p_{\text{s}} \propto p_s, \\
p_{\text{sca}} &\propto (p_s + p_i)\cdot r_{\text{togg}} + p_l, \\
p_{\text{all}} &\propto p_s + p_i + p_l.
\end{align}

% \begin{equation}
%     p_{\text{i}} \propto p_i, \qquad
%     p_{\text{s}} \propto p_s,
% \end{equation}
% \begin{equation}
%     p_{\text{sca}} \propto (p_s + p_i)\cdot r_{\text{togg}} + p_l,
% \end{equation}
% \begin{equation}
%     p_{\text{all}} \propto p_s + p_i + p_l.
% \end{equation}

% These quantities are spatially aggregated into uniform layout tiles. 
% The Internal Power and Switching power computed as:
% \begin{equation}
% p_{\text{i}} \propto  p_i 
% \end{equation}
% \begin{equation}
% p_{\text{s}} \propto  p_s 
% \end{equation}
% The toggle-scaled power is computed as:
% \begin{equation}
% p_{\text{sca}} \propto (p_s + p_i)\cdot r_{\text{togg}} + p_l,
% \end{equation}
% and the total power map is:
% \begin{equation}
% p_{\text{all}} \propto p_s + p_i + p_l.
% \end{equation}
% These instance level power is merged into corresponding tiles to create the power maps.

Timing window reports contains possible switching time domain of the instance in a clock period from a static timing analysis for each pin. The clock period is decomposed evenly into 20 parts, and the cell contributes to time-decomposed power map ($p_t$) only in the parts that it is switching.
\begin{equation}
p_t[k] \propto p_{\text{sca}}, \quad k = 0,\dots,19.
\end{equation}
These yield a $24$-channel feature image per layout region ($4$ static power components and $20$ temporal components).

% These IR drop features yields 24 channels feature image (each channel represent the IR drop feature in feature image). 

% While the above features capture how much current is drawn, they do not encode where switching activity is structurally concentrated nor how physical layout conditions influence localized IR drop. Consequently, methods\cite{chhabria2021mavirec, zhao2024pdnnetpdnawaregnncnnheterogeneous}  relying solely on power maps may struggle to differentiate regions with similar power magnitudes but different spatial connectivity or routing stress.

While power map based feature images capture current magnitude, they fail to encode structural locality and spatial context governing current paths and PDN resistance modulation. Regions with similar power intensity can exhibit substantially different IR-drop behavior depending on routing congestion, macro blockage, and pin clustering. Consequently, methods\cite{chhabria2021mavirec, zhao2024pdnnetpdnawaregnncnnheterogeneous} relying solely on power maps struggle to differentiate regions with similar power magnitudes but different spatial connectivity or routing stress.

% To address this limitation, we add geometry-aware features that complement the current proxies.
% Representative examples are visualized in Fig.~\ref{fig:risky_feature_ir}.
% \noindent \textbf{Additional layout-aware features.}
% To provide geometrical information, we add
% ten additional IR drop features (summarized in Table~\ref{tab:layout_features}). These features encode spatial factors that modulate either effective current density $I(x,y)$ or path resistance $R_{\text{eff}}(x,y)$.
% \noindent \textbf{Additional layout-aware features.}
% As described in the IR-drop formulation (\(Gv = i\)), the voltage drop at each tile depends jointly on the local current demand and the effective resistance of the PDN paths. 
% To better represent these spatial dependencies, we add ten additional layout-aware IR-drop features (summarized in Table~\ref{tab:layout_features}). 
% These features encode geometrical and routing factors that modulate either the effective current density $I(x,y)$ or the path resistance $R_{\text{eff}}(x,y)$, complementing the power-derived maps with structural and congestion-aware context.
\noindent \textbf{Additional layout-aware features.}
\begin{figure*}[t]
    \centering
    \includegraphics[width=0.98\linewidth]{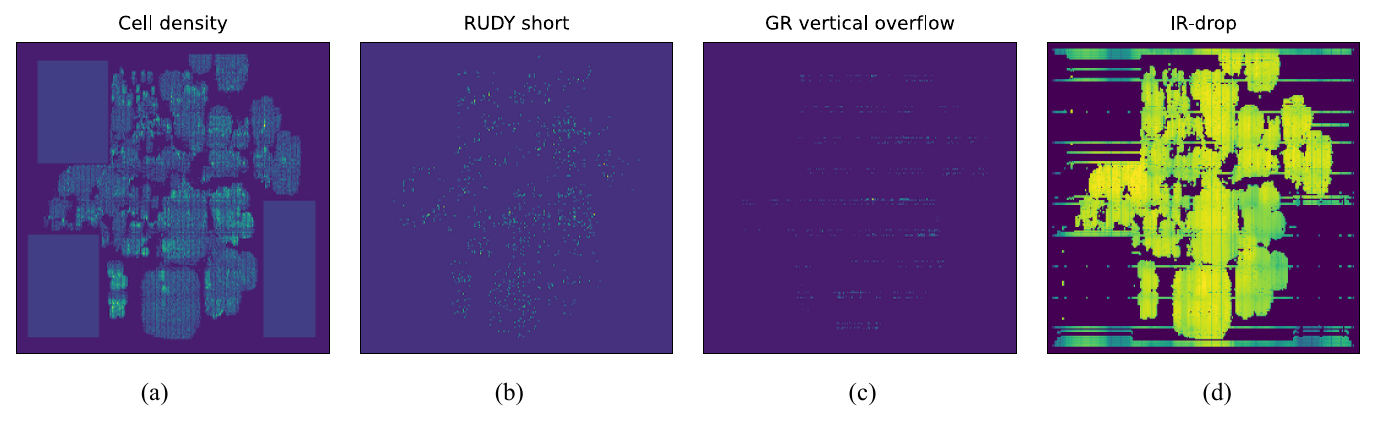}
    \caption{
        Visualization of additional features and ground-truth IR-drop map for the RISCY design from the N14 technology dataset.
        From left to right: (a) \textit{Cell Density}, (b) \textit{RUDY Short}, (c) \textit{Global Routing Vertical Overflow}, and (d) \textit{IR-drop Ground Truth}.  
    }
    \label{fig:risky_feature_ir}
\end{figure*}
As described in the IR-drop analysis (Eq.~\ref{eq:gv_i}), the voltage drop at each tile depends jointly on the local current demand and the effective resistance of the PDN paths. 
To better represent these spatial dependencies (geometric information), we add ten additional layout-aware IR-drop features (summarized in 
Table~\ref{tab:layout_features}). 
These features encode geometrical and routing factors that modulate either the effective current density $I_{\text{eff}}(x,y)$ or the path resistance $R_{\text{eff}}(x,y)$.

The $I_{\text{eff}}(x,y)$ current flowing through PDN network is given by:

% \hl{are these notations, RUDY, cell\_density, macro\_region standard or also used in prior works? Use one letter optionally with subscript will look better}

\begin{equation}
\label{eq:iproxy_final}
\begin{aligned}
I_{\mathrm{eff}}(x,y) &= 
\frac{P(x,y)}{V_{dd}}
+ \alpha_{1} R(x,y)
+ \alpha_{2} R_{\mathrm{pin}}(x,y) \\
&\quad
+ \alpha_{3} D(x,y)
+ \alpha_{4} M(x,y),
\end{aligned}
\end{equation}
where $P(x,y)$ is the total aggregated power in tile $(x,y)$ from power maps; 
$R(x,y)$ is the RUDY (Rectangular Uniform wire Density); 
$R_{\mathrm{pin}}(x,y)$ is the pin-RUDY map capturing pin-driven routing density; 
$D(x,y)$ is the cell-density map representing standard-cell counts per tile; 
$M(x,y)$ is the macro-region indicator identifying tiles occupied by macros; 
and $\alpha_i \ge 0$ are scaling coefficients that weight the contribution of each geometric feature.

% \begin{equation}
% \label{eq:Ieff_final}
% \begin{aligned}
% I_{\mathrm{eff}}(x,y) &= 
% \frac{p(x,y)}{V_{dd}}
% + \alpha_1\, r(x,y) 
% + \alpha_2\, r_p(x,y) 
% + \alpha_3\, d(x,y) \\
% &\quad
% + \alpha_4\, m(x,y)
% + \alpha_5\, r_\ell(x,y)
% + \alpha_6\, r_s(x,y),
% \end{aligned}
% \end{equation}

% \begin{equation}
% \label{eq:iproxy_clean}
% \begin{aligned}
% I_{\mathrm{eff}}(x,y) &= 
% \frac{P(x,y)}{V_{dd}}
% + \alpha_1 R(x,y)
% + \alpha_2 R_{\mathrm{pin}}(x,y) \\
% &\quad
% + \alpha_3 D(x,y)
% + \alpha_4 M(x,y),
% \end{aligned}
% \end{equation}

% \begin{equation}
% \begin{split}
% I_{\text{eff}}(x,y) &= \frac{p_{\text{all}}(x,y)}{V_{dd}} 
% + \alpha_1\,\text{RUDY}(x,y)\\&
% + \alpha_2\,\text{RUDY}_{\text{pin}}(x,y) 
%  + \alpha_3\,\text{cell\_density}(x,y) \\&
% \quad+ \alpha_4\,\text{macro\_region}(x,y)
% \end{split}
% \label{eq:iproxy}
% \end{equation}
% where $\alpha_i \ge 0$ are coefficients that adaptively scale the spatial influence of each feature.  
\begin{equation}
\label{eq:reff_final}
\begin{aligned}
R_{\mathrm{eff}}(x,y) &=
R_{0}(x,y)\Bigl[
1 
+ \beta_{1} O^{\mathrm{eGR}}_{H}(x,y)
+ \beta_{2} O^{\mathrm{eGR}}_{V}(x,y) \\
&\qquad
+ \beta_{3} O^{\mathrm{GR}}_{H}(x,y)
+ \beta_{4} O^{\mathrm{GR}}_{V}(x,y)
\Bigr],
\end{aligned}
\end{equation}
where $R_{0}(x,y)$ is the nominal PDN resistance in tile $(x,y)$; 
$O^{\mathrm{eGR}}_{H}(x,y)$ and $O^{\mathrm{eGR}}_{V}(x,y)$ denote early-global-routing (eGR) horizontal and vertical overflow maps; 
$O^{\mathrm{GR}}_{H}(x,y)$ and $O^{\mathrm{GR}}_{V}(x,y)$ denote global-routing (GR) horizontal and vertical overflow maps; 
and $\beta_{j} \ge 0$ are coefficients that modulate the influence of directional routing congestion on the effective resistance.
% % The routing overflow maps provide weak priors on effective local resistance:
% \begin{equation}
% \begin{split}
% R_{\text{eff}}(x,y) &\approx R_0(x,y)\Bigl[
% 1 + \beta_1\,\text{eGR\_H}(x,y) \\
% &+ \beta_2\,\text{eGR\_V}(x,y) 
% + \beta_3\,\text{GR\_H}(x,y)\\
% &\qquad + \beta_4\,\text{GR\_V}(x,y)
% \Bigr]
% \end{split}
% \label{eq:reff}
% \end{equation}
% where $R_0(x,y)$ denotes the nominal PDN resistance at tile $(x,y)$ and $\beta_j \ge 0$ are scaling factors.  
Together, these relations form a physically consistent approximation:
\begin{equation}
V_{\text{drop}}(x,y) \approx R_{\text{eff}}(x,y)\,I_{\text{eff}}(x,y).
\end{equation}

% \begin{table}[t]
% \centering
% \caption{Layout-aware feature groups fused with power maps and their contribution to the current proxy $I_{\mathrm{eff}}$ and the resistance proxy $R_{\mathrm{eff}}$.}
% \small
% \begin{tabular}{l l c c}
% \toprule
% \textbf{Feature Group} & \textbf{Channels} & \textbf{$I_{\mathrm{eff}}$} & \textbf{$R_{\mathrm{eff}}$} \\
% \midrule

% Cell density &
% $\text{C}_{\text{den}}(x,y)$ &
% \checkmark & \xmark \\[4pt]

% Macro region &
% $\text{M}(x,y)$ &
% \checkmark & \xmark \\[6pt]

% \multirow{2}{*}{RUDY-based demand} &
% $\text{RUDY},\; \text{RUDY}_{\text{pin}}$ &
% \checkmark & \xmark \\[2pt]
% & $\text{RUDY}_{\text{long}},\; \text{RUDY}_{\text{short}}$ &
% \checkmark & \xmark \\[6pt]

% \multirow{2}{*}{Routing overflow} &
% $\text{O}^{\mathrm{eGR}}_{H},\; \text{O}^{\mathrm{eGR}}_{V}$ &
% \xmark & \checkmark \\[2pt]
% & $\text{O}^{\mathrm{GR}}_{H},\; \text{O}^{\mathrm{GR}}_{V}$ &
% \xmark & \checkmark \\
% \bottomrule
% \end{tabular}
% \label{tab:layout_features}
% \end{table}

\begin{table}[t]
\centering
\caption{Layout-aware feature groups fused with power maps and their contribution to the current proxy $I_{\mathrm{eff}}$ and the resistance proxy $R_{\mathrm{eff}}$.}
\small
\setlength{\tabcolsep}{7pt}
\begin{tabular}{l l c c}
\toprule
\textbf{Feature Group} & \textbf{Channels} & \textbf{$I_{\mathrm{eff}}$} & \textbf{$R_{\mathrm{eff}}$} \\
\midrule

Cell density &
$\text{C}_{\text{den}}(x,y)$ &
\checkmark & \xmark \\

Macro region &
$\text{M}(x,y)$ &
\checkmark & \xmark \\

RUDY-based demand &
$\text{RUDY},\; \text{RUDY}_{\text{pin}}$ &
\checkmark & \xmark \\

RUDY-based demand &
$\text{RUDY}_{\text{long}},\; \text{RUDY}_{\text{short}}$ &
\checkmark & \xmark \\

Routing overflow &
$\text{O}^{\mathrm{eGR}}_{H},\; \text{O}^{\mathrm{eGR}}_{V}$ &
\xmark & \checkmark \\

Routing overflow &
$\text{O}^{\mathrm{GR}}_{H},\; \text{O}^{\mathrm{GR}}_{V}$ &
\xmark & \checkmark \\

\bottomrule
\end{tabular}
\label{tab:layout_features}
\end{table}

% The additional channels do not directly encode PDN geometry but provide contextual cues that influence effective current flow and resistance variation.
This enables the feature image to capture both local and long-range spatial dependencies. Including these ten layout-aware features yields a $256\times256\times34$ feature tensor feature image, significantly enhancing the generative diffusion process by embedding physical-design structure and routability priors into the learned IR-drop representation. The Figure \ref{fig:risky_feature_ir} shows additional feature visualization (three shown) for a RISCY  design. Three features are cell density, RUDY short, and GR vertical overflow.
% \textcolor{blue}{Kiran: add citation}
\subsection{Graph Creation Based on Topological Information}
\begin{figure}[t]
  \centering
  \includegraphics[width=0.95\linewidth]{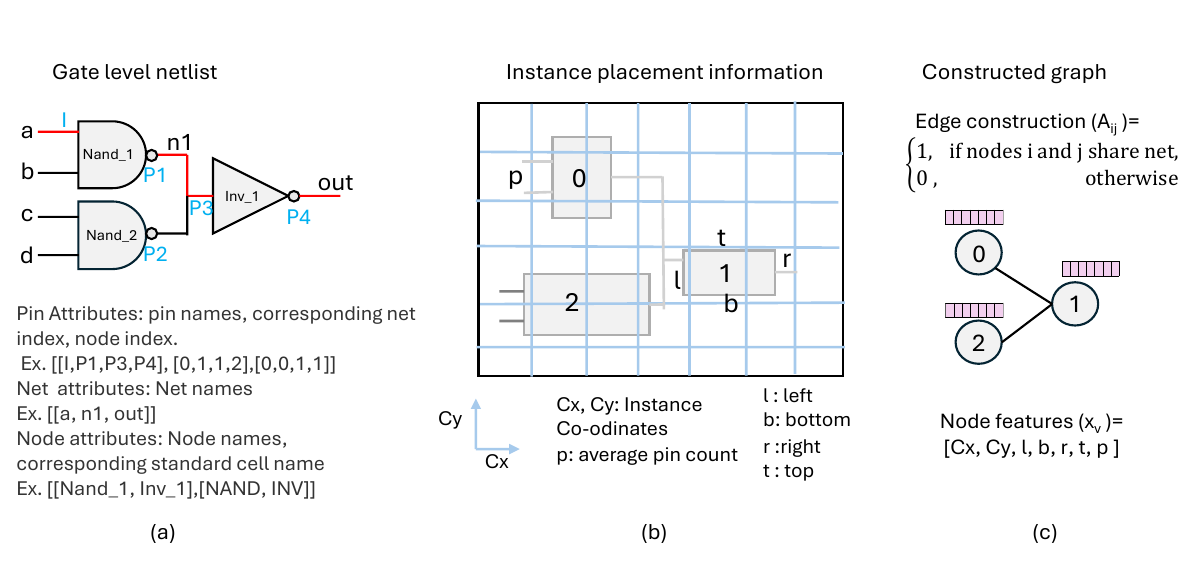}
  \caption{Graph construction: (a) Gate level netlist and graph construction attributes, (b) Instance (GCell) placement information, each instance is placed on grid  $(c_x,c_y)$, annotated with its bounding‑box coordinates $(l,b,r,t)$, and pin count $p$, (c) Constructed graph representation with node feature vector $\mathbf{x}_v = [c_x,\, c_y,\, l,\, b,\, r,\, t,\, p]$.}
  \label{fig:node_mapping}
\end{figure}

To incorporate logical connectivity into IR-drop generation, we construct a graph for each design using the flattened gate-level netlist and instance placement data provided in CircuitNet \cite{chai2023circuitnet,jiang2024circuitnet}. Each design is represented as $\mathcal{G} = (\mathcal{V}, \mathcal{E}, \mathbf{X})$, where $\mathcal{V}$ is the set of instances, $\mathcal{E}$ encodes connectivity through nets, and $\mathbf{X}$ contains physical design node features.
% The dataset provides three arrays: (1) \emph{pin attributes} (pin name, net index, node index), (2) \emph{net attributes} (net names), and (3) \emph{node attributes} (instance names and cell types).
% A example is illustrated in Fig.~\ref{fig:node_mapping}(a).
Pins that share the same \emph{net index} belong to the same net; pins that share the same \emph{node index} belong to the same instance. For example, in Figure~\ref{fig:node_mapping}(a), pins P1 and P3 both have net index 1, so they belong to net \texttt{n1}, which connects \texttt{NAND\_1} and \texttt{INV\_1}. By traversing the pin attributes, we identify all nets and the set of instances connected through each net.

Two instances are connected by an edge if they appear together on at least one net:
\begin{equation}
\mathbf{A}_{ij} = 
\begin{cases}
1, & \text{if instances } i \text{ and } j \text{ share a net},\\[-0.2em]
0, & \text{otherwise}.
\end{cases}
\end{equation}

The example in Figure~\ref{fig:node_mapping}(a) shows how the three attribute arrays describe the circuit. The pin-attribute array lists the pin names together with their net indices and node indices (e.g., \texttt{I}, \texttt{P1}, \texttt{P3}, \texttt{P4} belong to nets \{0,1,1,2\} and nodes \{0,0,1,1\}). The net-attribute array simply maps each net index to its net name (e.g., net~0 corresponds to \texttt{a}, net~1 to \texttt{n1}, and net~2 to \texttt{out}). The node-attribute array maps each node index to an instance and its standard-cell type (e.g., node~0 is \texttt{NAND\_1} of type \texttt{NAND}, and node~1 is \texttt{INV\_1} of type \texttt{INV}). Together, these three arrays fully specify which pins belong to each net and which pins belong to each instance, allowing the logical connectivity to be reconstructed directly from the attributes.
Placement information is provided as bounding-box coordinates for each instance in GCell units (Figure~\ref{fig:node_mapping}(b)). For each node $v$, we construct the feature vector
\begin{equation}
\mathbf{x}_v = [c_x,\,c_y,\,l,\,b,\,r,\,t,\,p],
\end{equation}
where $(c_x,c_y)$ is the instance center, $(l,b,r,t)$ are the bounding-box coordinates, and $p$ is the pin count. 
The final constructed graph is shown in Figure~\ref{fig:node_mapping}(c). 
% Figure~\ref{fig:Graph_constrcuted}.
Each node encodes geometric layout properties, and each edge reflects logical connectivity derived from the netlist. All graphs are stored in PyTorch Geometric format, enabling efficient integration into our multimodal generative framework.

\subsection{Image--Graph Fusion for Multimodal Conditioning of Diffusion Model}
\label{sec:fusion}

Figure~\ref{fig:overview} (c) illustrates how GIF fuses geometric layout features and logical netlist topology inside a diffusion model tailored to IR-drop generation. A detailed illustration of the fusion mechanism is provided in the supplementary material (Image Graph Fusion Mechanism). Although diffusion models and UNet-based conditional parameterizations are widely used in vision~\cite{ho2020denoising,rombach2022high}, their standard conditioning interfaces do not reflect the mixed local–global dependencies that govern IR-drop behavior. Our approach is not to introduce a new generative mechanism, but to construct a conditioning pathway that matches the physical factors underlying IR-drop: localized current demand shaped by geometry, and long-range correlations shaped by netlist topology.

Let $Y \in \mathbb{R}^{H \times W}$ denote a ground truth IR-drop field. Following the implementation, we normalize IR-drop values into the range $[-1,1]$ via $\widetilde{Y} = 2Y - 1$. At diffusion step $t$, the model observes a noisy version
\[
\widetilde{Y}_t
= \sqrt{\bar{\alpha}_t}\, \widetilde{Y}
  + \sqrt{1 - \bar{\alpha}_t}\, \varepsilon,
\qquad
\varepsilon \sim \mathcal{N}(0, \mathbf{I}_{H \times W}),
\]
which reflects a physically plausible IR-drop field perturbed by uncertainty in local current demand. The diffusion model learns to predict the noise component $\varepsilon$ conditioned on both geometric layout features and netlist topology.

Each tile is represented by a multi-channel geometric feature map 
$X \in \mathbb{R}^{C \times H \times W}$ encoding power maps and additional features from layout. Because IR-drop varies sharply with these local factors, $X$ provides spatial conditioning throughout the UNet. Consistent with FiLM-style feature modulation~\cite{perez2018film} and adaptive normalization techniques such as AdaIN/AdaGN~\cite{huang2017arbitrary}, each intermediate UNet feature map $F^{(\ell)} \in \mathbb{R}^{C_\ell \times H_\ell \times W_\ell}$ is modulated by a pair of learned affine transforms depending on the downsampled layout features $X_\ell$ and the timestep embedding:
\[
\widehat{F}^{(\ell)}
    = \gamma^{(\ell)}(X_\ell, t) \odot F^{(\ell)}
      + \beta^{(\ell)}(X_\ell, t).
\]

This matches the implementation, where time- and geometry-conditioned scales and shifts are produced by small MLPs and applied multiplicatively and additively inside each ResBlock.  
This mechanism encourages the diffusion model to encode fine-grained IR-drop structure tied to spatial variations in current demand.
% Local geometry alone, however, does not determine IR-drop. The circuit’s logical netlist induces long-range current-demand correlations through shared drivers, sinks, and power paths.
To incorporate topology information, the netlist graph $G=(V,E)$ is processed by a lightweight GCN encoder,
\[
H = \phi_{\mathrm{gcn}}(G) \in \mathbb{R}^{N \times D},
\]
following standard message-passing formulations. Because designs vary widely in size, we compress the node embeddings $H$ into a fixed set of $K$ topology tokens
\[
T = \rho(H) \in \mathbb{R}^{K \times D},
\]
using a permutation-invariant pooling operator (degree-aware or mean pooling). These tokens summarize global logical structure at tile granularity without introducing design-size–dependent computational cost.

To allow global topology to influence IR-drop generation, the tokens $T$ are injected into the UNet through cross-attention layers inspired by multimodal conditioning in latent diffusion~\cite{rombach2022high}.
At a low-resolution layer $\ell$ corresponding to the spatial scale where IR-drop exhibits large spatial smoothness and global coupling. We treat the spatial features as queries and the topology tokens as keys and values:
\[
\Delta F^{(\ell)} 
= 
\mathrm{softmax}\!\left(
    \frac{Q^{(\ell)} K^\top}{\sqrt{D_q}}
  \right) V,
\]
where $Q^{(\ell)}$ is derived from $\widehat{F}^{(\ell)}$ and $K,V$ are linear projections of $T$. This matches the implementation: the feature map is flattened into $(H_\ell W_\ell)$ queries, projected into multi-head format, and fused with token-derived values.
The topology-aware update is incorporated using a learnable scalar gate $\alpha_\ell$ initialized to zero, exactly as implemented:
\[
F_{\text{fused}}^{(\ell)}
    = \widehat{F}^{(\ell)}
    + \tanh(\alpha_\ell)\,
      \mathrm{reshape}\!\left(\Delta F^{(\ell)}\right).
\]

This guarantees that the model initially behaves as a purely geometry conditioned denoiser and progressively learns to incorporate topology only if it improves IR-drop generation. The use of gated cross-attention is not intended as a new attention mechanism; it is a domain-specific choice motivated by how global logical connectivity influences low-frequency IR-drop structure.
During sampling, the reverse diffusion process uses this fused UNet at every timestep, conditioned jointly on the geometric map $X$ and the topology tokens $T$.
% The reverse update follows the standard DDPM mean–variance computation, while the denoising network itself embeds the physically motivated fusion of local geometry and global topology described above.
This design aligns the generative model with the underlying mechanisms of IR-drop formation, enabling GIF to generate IR-drop maps consistent with both spatial layout and logical circuit structure.

\section{Evaluation}
\label{sec:eval}
\noindent \textbf{Experimental Setup.}  
All experiments use Ubuntu~22.04 with an AMD~EPYC 7763 CPU (64~cores, 128~threads) and four NVIDIA~RTX~A6000 GPUs (48\,GB memory each, CUDA~12.6, driver~560.35.05).

\noindent \textbf{Model and Training Parameters.}
We adopt a U-Net-based~\cite{ronneberger2015unetconvolutionalnetworksbiomedical} denoising backbone with $T{=}1000$ diffusion steps. 
We use AdamW ($\beta_1{=}0.9$, $\beta_2{=}0.999$) with learning rate $2\times10^{-4}$ and no weight decay. 
An auxiliary reconstruction loss (L1 on $\hat{x}_0$) with weight $0.1$ is applied only for diffusion steps $t<150$. 
Exponential moving average (EMA) weights with decay $0.999$ are used for evaluation.
% \noindent \textbf{Model and Training Parameters.}
% We adopt a UNet-based~\cite{ronneberger2015unetconvolutionalnetworksbiomedical} denoising backbone with $T{=}1000$ diffusion steps using a cosine $\beta$ schedule.    
% Training is performed for $60$~epochs with a batch size of~64 using the Adam optimizer ($\beta_1{=}0.9$, $\beta_2{=}0.999$), learning rate $2\times10^{-4}$, and no weight decay.   
% An auxiliary loss with weight~0.1 is applied for diffusion steps $t<150$.  
% Exponential moving average (EMA) weights with decay~0.999 are employed for evaluation.  
\subsection{Datasets}
We evaluate our framework on the CircuitNet-N28 and CircuitNet-N14 datasets~\cite{jiang2024circuitnet,chai2023circuitnet}. Our formulation operates on spatial layout representations and is therefore applicable across technology nodes. Evaluating on both 28\,nm and 14\,nm technology designs allows us to assess the ability of the model to generalize across different fabrication technologies and layout characteristics. Additional dataset details are provided in the supplementary material (Dataset Details).

\noindent \textbf{CircuitNet-N28.}
CircuitNet-N28 contains physical design features and IR-drop maps generated from six open-source RISC-V designs fabricated in 28\,nm technology. We adopt a design-wise split with four designs for training, one for validation, and one for testing to prevent cross-design leakage (Table~\ref{tab:dataset_split}).

\noindent \textbf{CircuitNet-N14.}
CircuitNet-N14 includes a broader set of designs such as RISC-V processors, NVIDIA GPUs, and ML accelerators fabricated in 14\,nm technology. Compared to N28, these layouts exhibit greater variation in floorplan organization, utilization, aspect ratios, and power delivery configurations.
We again apply a design-wise split, using six designs for training, Vortex-small for validation, and zero-riscy for testing (Table~\ref{tab:dataset_split_N14}).

% For consistent training across both technology nodes, all feature maps and IR-drop labels are resized to $256\times256$ and normalized before being used by the model.

% \subsection{Datasets} 
% We evaluate our framework on both the CircuitNet-N28 and CircuitNet-N14 datasets \cite{jiang2024circuitnet,chai2023circuitnet}. 
% CircuitNet-N28 is constructed using six open-source RISC-V (Reduced Instruction Set Computer architecture) designs fabricated in 28\,nm planar technology. 
% It contains 10{,}242 samples, each consisting of a 24-channel feature image and the corresponding IR-drop label map.  CircuitNet-N14 includes a broader mix of designs, including RISC-V, GPU, and ML accelerators, fabricated in 14\,nm FinFET technology. It contains  {10{,}444} samples, each consisting of a 24-channel feature image and the corresponding IR-drop label map.   To avoid cross-design leakage, we use a design-wise split in all experiments.
\begin{table}[t]
\centering
\begin{minipage}[t]{0.48\columnwidth}
\centering
\caption{Design-wise split of the CircuitNet-N28 dataset.}
\label{tab:dataset_split}
\small
\setlength{\tabcolsep}{6pt}
\begin{tabular}{lcc}
\toprule
\textbf{Design} & \textbf{Split} & \textbf{Samples} \\
\midrule
RISCY-a        & Train & 2{,}003 \\
RISCY-b        & Train & 1{,}858 \\
RISCY-FPU-a    & Train & 1{,}969 \\
zero-riscy-a   & Train & 2{,}042 \\
RISCY-FPU-b    & Val   & 1{,}248 \\
\midrule
\textbf{zero-riscy-b}   & \textbf{Test}  & \textbf{1{,}122} \\
\midrule
\textbf{Total} &       & \textbf{10{,}242} \\
\bottomrule
\end{tabular}
\end{minipage}
\hfill
\begin{minipage}[t]{0.48\columnwidth}
\centering
\caption{Design-wise split of the CircuitNet-N14 dataset.}
\label{tab:dataset_split_N14}
\setlength{\tabcolsep}{6pt}
\small
\begin{tabular}{lcc}
\toprule
\textbf{Design} & \textbf{Split} &   \textbf{Samples} \\
\midrule
Nvidia-small    & Train      & 85 \\
RISCY          & Train      & 3{,}162 \\
RISCY-FPU      & Train      & 3{,}456 \\
Vortex-large   & Train      & 61 \\
openc910-1     & Train      & 96 \\
Nvidia-large    & Train      & 32 \\
Vortex-small   & Val & 96 \\
\midrule
\textbf{zero-riscy}     & \textbf{Test}      & \textbf{3{,}456} \\
\midrule
\textbf{Total} &            & \textbf{10{,}444} \\
\bottomrule
\end{tabular}
\end{minipage}
\end{table}
\subsection{Evaluation Metrics}
We assess generated IR-drop maps by comparing them against ground truth IR drop maps using metrics commonly adopted in prior IR-drop modeling work \cite{Xie_2020, chhabria2021mavirec, zhao2024pdnnetpdnawaregnncnnheterogeneous}. 
PSNR measures per-pixel distortion, while SSIM reflects similarity in local spatial structure and contrast \cite{fardo2016formal, 1284395}. 
MAE and RMSE quantify the average and large-magnitude differences between generated and reference IR-drop values. 
Pearson and Spearman correlations evaluate how well the generated maps preserve the overall IR-drop distribution and the relative severity ordering across layout regions. 
Together, these metrics characterize pixel-wise accuracy, structural similarity, and global trend consistency of the generated IR-drop maps. Further discussion of these metrics in the context of IR-drop analysis is provided in the supplementary material (Evaluation Metrics).

\subsection{IR-drop Map Generation Evaluation}
We evaluate GIF on the CircuitNet-N28  dataset using PSNR, SSIM, MAE, RMSE, Pearson, and Spearman metrics (Table~\ref{tab:n28_results}). These metrics measure pixel accuracy and structural consistency.
Adding the geometric feature representation (24 to 34 channels) increases PSNR from 19.218 to 19.542 and SSIM from 0.699 to 0.739. Adding ControlNet further increases PSNR to 19.583 and SSIM to 0.766, and also lowers MAE and RMSE.
Adding topological information through graph-conditioned cross-attention provides additional gains. The $K=32$ top-$k$ configuration (without ControlNet) increases Pearson correlation to 0.9130 while keeping PSNR at a similar level. The $K=64$ mean-pooled configuration with ControlNet achieves the highest SSIM (0.785) and the highest correlations (Pearson 0.9215, Spearman 0.5139), with a small decrease in PSNR to 18.935.
Overall, the image-only configuration with ControlNet gives the highest pixel accuracy, while the graph-conditioned configurations give higher structural consistency and higher correlation with the true IR-drop maps.
\begin{table*}[t]
\centering
%\small
\caption{
Quantitative generation results on the CircuitNet-N28 \cite{jiang2024circuitnet,chai2023circuitnet} dataset. 
}
\resizebox{\textwidth}{!}{
\begin{tabular}{lccccccc}
\toprule
\textbf{Conditioning} & \textbf{ControlNet} & \textbf{PSNR}↑ & \textbf{SSIM}↑ & \textbf{MAE}↓ & \textbf{RMSE}↓ & \textbf{Pearson}↑ & \textbf{Spearman}↑ \\
\midrule
24-ch images & \xmark & 19.218 & 0.699 & 0.0378 & 0.1103 & 0.9069 & 0.4949 \\
34-ch images & \xmark & 19.542 & 0.739 & 0.0352 & 0.1063 & 0.9118 & 0.4973 \\
24-ch images & \cmark & 19.255 & 0.711 & 0.0374 & 0.1098 & 0.9065 & 0.4865 \\

34-ch images & \cmark & \textbf{19.583} & 0.766 & \textbf{0.0345} & \textbf{0.1058} & 0.9127 & 0.5033 \\
34-ch + graph (K=32, topk) & \xmark & 19.519 & 0.611 & 0.0383 & 0.1066 & 0.9130 & 0.5107 \\
34-ch + graph (K=32, topk) & \cmark & 19.307 & 0.762 & 0.0363 & 0.1091 & 0.9064 & 0.4976 \\
34-ch + graph (K=64, mean) & \cmark & 18.935 & \textbf{0.785} & 0.0405 & 0.1141 & \textbf{0.9215} & \textbf{0.5139} \\
\bottomrule
\end{tabular}
}
\label{tab:n28_results}
\end{table*}

% \subsubsection{Comparison with Existing Methods}
\noindent \textbf{Comparison with Existing Methods.}
For fair comparison, we follow the same evaluation protocol and metrics used in prior work. 
All experiments use the CircuitNet-N28 dataset under the design split defined by PDNNet. 
Four designs (\textit{RISCY-a}, \textit{RISCY-b}, \textit{RISCY-FPU-a}, \textit{RISCY-FPU-b}) are used for training, 
and two unseen designs (\textit{zero-riscy-a}, \textit{zero-riscy-b}) are used for testing. 
The model uses the AdamW optimizer with a learning rate of $8{\times}10^{-4}$, 
EMA decay of 0.999, and a cosine learning-rate schedule. 
The diffusion process uses 1000 cosine-scheduled steps. 
GIF uses a conditional diffusion model with 34-channel layout-aware inputs, 
graph tokens extracted from the netlist, and cross-attention for joint conditioning.
Table~\ref{tab:sota_comparison} reports the results. 
Baseline results for PowerNet~\cite{Xie_2020}, MAVIREC~\cite{chhabria2021mavirec}, 
and PDNNet~\cite{zhao2024pdnnetpdnawaregnncnnheterogeneous} are taken from their published performance on the same dataset. 
GIF achieves the lowest NMAE (0.0266) and increases PSNR, SSIM, and Pearson correlation to 21.77\,dB, 0.786, and 0.9536, respectively. 
Overall, GIF reaches higher pixel accuracy and higher structural consistency than prior methods across all evaluation metrics.

%TODO 

% \textcolor{red}{ Mostafa: make sure that the tabel font size matches with the text}
% \begin{table}[t]
% \centering
% \small
% %\scriptsize
% % \setlength{\tabcolsep}{ 8 pt}
% \caption{Comparison with existing methods on CircuitNet-N28}
% \label{tab:sota_comparison}
% \begin{tabular}{lcccc}
% \toprule
% \textbf{Method} & \textbf{NMAE} $\downarrow$ & \textbf{PSNR} $\uparrow$ & 
% \textbf{SSIM} $\uparrow$ & \textbf{Pear} $\uparrow$ \\
% \midrule
% PowerNet~\cite{Xie_2020} & 0.149 & 11.60 & 0.56 & 0.77 \\
% MAVIREC~\cite{chhabria2021mavirec} & 0.039 & 18.27 & 0.68 & 0.91 \\
% PDNNet~\cite{zhao2024pdnnetpdnawaregnncnnheterogeneous} & 0.028 & 19.35 & 0.72 & 0.92 \\
% \textbf{Ours (GIF)} & \textbf{0.026} & \textbf{21.77} & \textbf{0.78} & \textbf{0.953} \\
% \bottomrule
% \end{tabular}
% \end{table}

\begin{table}[t]
\centering
\caption{Comparison with SOTA methods on CircuitNet-N28}
\label{tab:sota_comparison}
\small
\setlength{\tabcolsep}{6pt}
\begin{tabular}{lcccc}
\toprule
\textbf{Method} & \textbf{NMAE} $\downarrow$ & \textbf{PSNR} $\uparrow$ & 
\textbf{SSIM} $\uparrow$ & \textbf{Pear} $\uparrow$ \\
\midrule
PowerNet~\cite{Xie_2020} & 0.149 & 11.60 & 0.56 & 0.77 \\
MAVIREC~\cite{chhabria2021mavirec} & 0.039 & 18.27 & 0.68 & 0.91 \\
PDNNet~\cite{zhao2024pdnnetpdnawaregnncnnheterogeneous} & 0.028 & 19.35 & 0.72 & 0.92 \\
\textbf{Ours (GIF)} & \textbf{0.026} & \textbf{21.77} & \textbf{0.78} & \textbf{0.95} \\
\bottomrule
\end{tabular}
\end{table}

\begin{figure*}[t]
    \centering
    \includegraphics[width=0.95\linewidth]{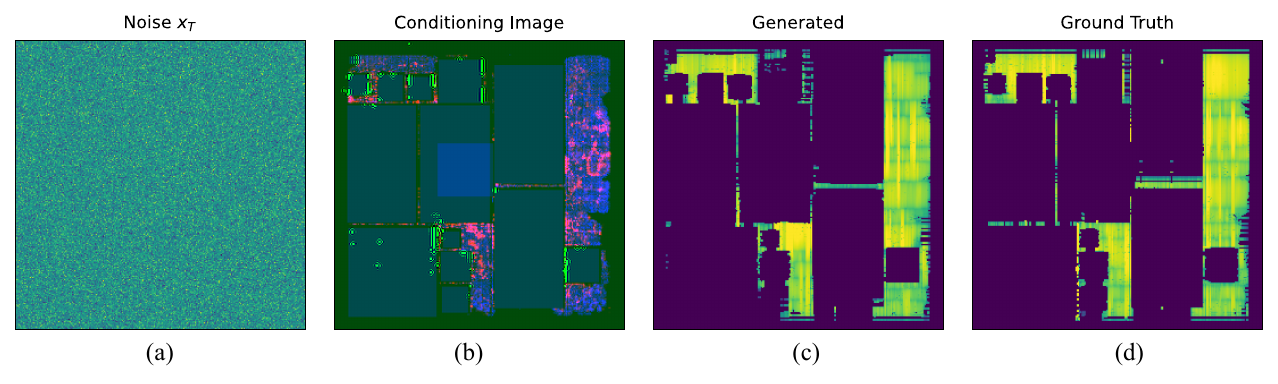}
 \caption{Qualitative IR-drop generation on CircuitNet-N28: 
(a) noise $x_T$, 
(b) conditioning features (3-channels shown), 
(c) generated IR-drop $\hat{x}_0$, 
(d) ground truth.}
 \label{fig:qualitative_n28}
\end{figure*}

% \subsubsection{Qualitative Visualization}
\noindent \textbf{Qualitative Visualization.}
\label{subsec:qualitative}
To complement the quantitative evaluation, we provide a representative CircuitNet-N28  test instance generated by our best graph-conditioned model (34-channel geometric features with $K{=}64$ mean-pooled graph tokens and ControlNet); see Figure~\ref{fig:qualitative_n28}. 
From left to right, the figure shows the initial Gaussian noise $x_T$, a three-channel composite of selected conditioning features, the generated IR-drop map $\hat{x}_0$, and the ground-truth IR-drop map. 
Because the conditioning tensor has 34 channels, we display only three representative feature maps as an RGB composite to provide a compact visualization; showing all channels is not visually interpretable.
For the sample design zero-riscy (zero-riscy-b-3-c2-u0.85-m1-p6-f1), the model outputs a PSNR of 19.625, an SSIM of 0.811, an MAE of 0.0333, an RMSE of 0.1044, a Pearson correlation of 0.9320, and a Spearman correlation of 0.5039. 
These values align with the aggregate results in Table~\ref{tab:n28_results}. 
In the visualization, the generated IR-drop map follows the overall magnitude and spatial distribution of the ground truth, including the primary high-drop regions and surrounding local variations.

\noindent \textbf{IR-drop Map Generation Evaluation on CircuitNet-N14.}
Table~\ref{tab:n14_results} summarizes the results on CircuitNet-N14. 
To the best of our knowledge, GIF is the first work to report IR-drop results on CircuitNet-N14, establishing initial baselines for future methods.
GIF shows strong generation ability, with Pearson correlation up to 0.9106 and Spearman correlation up to 0.8284. 
These values indicate that the model follows the overall IR-drop trends across the layout, even under the higher variability of the N14 dataset.
The model trained with a classifier-free dropout of 0.1 gives the best overall performance. 
It achieves the highest correlations (Pearson 0.9106, Spearman 0.8284), the lowest MAE (0.0667) and RMSE (0.1797), and the highest PSNR (14.987) and SSIM (0.558).
Other configurations, including the graph-conditioned variants, show lower PSNR and SSIM. 
Further analysis of the CircuitNet-N14 dataset and graph availability is provided in the supplementary material~\ref{sec:supp_additional_analysis}.
% (Additional Analysis on CircuitNet-N14 Generation). 
This behavior is expected because the N14 dataset contains larger variation in design styles and an imbalanced sample distribution, which makes pixel-based metrics more difficult to optimize.
% Other configurations, including the graph-conditioned variants, show lower PSNR and SSIM. 
% This behavior is expected because the N14 dataset contains larger variation in design styles, an imbalanced sample distribution, which makes pixel-based metrics more difficult to optimize.
Overall, GIF maintains high correlation accuracy across both N28 and N14, indicating that the framework generalizes across different technology nodes.

\begin{table}[t]
\centering
\caption{Quantitative results on the CircuitNet-N14 dataset.}
\label{tab:n14_results}
%\small
\resizebox{\columnwidth}{!}{%
\begin{tabular}{lcccccc}
\toprule
\textbf{Model} & \textbf{PSNR}↑ & \textbf{SSIM}↑ & \textbf{MAE}↓ & \textbf{RMSE}↓ & \textbf{Pearson}↑ & \textbf{Spearman}↑ \\
\midrule
34-ch + ControlNet (no CFG) 
& 14.156 & 0.513 & 0.0795 & 0.1996 & 0.8903 & 0.8133 \\

34-ch + ControlNet (cfg-drop 0.1, no CFG at test) 
& \textbf{14.987} & \textbf{0.558} & \textbf{0.0667} & \textbf{0.1797} & \textbf{0.9106} & \textbf{0.8284} \\

34-ch + Graph (K=32, top-k) + ControlNet (no CFG) 
& 14.143 & 0.4947 & 0.0797 & 0.1995 & 0.8892 & 0.8103 \\

34-ch + Graph (K=32, top-k) + ControlNet (finetuned, no CFG) 
& 14.455 & 0.5025 & 0.0758 & 0.1918 & 0.8974 & 0.8149 \\
\bottomrule
\end{tabular}
}
\end{table}
% \subsubsection{Qualitative Visualization on CircuitNet-N14}
\noindent \textbf{Qualitative Visualization on CircuitNet-N14.}
\begin{figure*}[!t]
    \centering
    \includegraphics[width=0.99\linewidth]{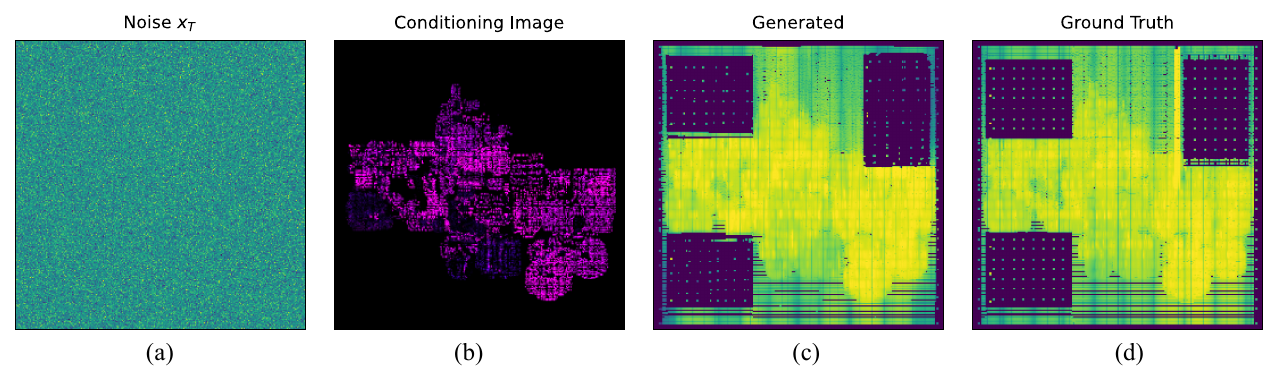}
     \caption{Qualitative IR-drop generation on CircuitNet-N14: 
(a) noise $x_T$, 
(b) conditioning features (3-channels shown), 
(c) generated IR-drop $\hat{x}_0$, 
(d) ground truth.}
    \label{fig:qualitative_n14}
\end{figure*}
To complement the quantitative evaluation, we provide a representative N14 test instance generated by our image-only model (34-channel conditioning with ControlNet and a classifier-free dropout rate of 0.1); see Figure~\ref{fig:qualitative_n14}. 
As in the N28 visualization, the figure shows the initial Gaussian noise $x_T$, a three-channel composite of selected conditioning features, the generated IR-drop map $\hat{x}_0$, and the ground-truth IR-drop map. 
% Because the conditioning tensor has 34 channels, we display only three representative channels as an RGB composite to offer an interpretable summary.
For the shown zero-riscy sample, the model yields a PSNR of 16.818, an SSIM of 0.518, an MAE of 0.0593, an RMSE of 0.1442, a Pearson correlation of 0.9243, and a Spearman correlation of 0.8922. 
These values are consistent with the overall N14 trends, where correlation metrics remain strong despite the larger design variability and the reduced spatial regularity of 14\,nm layouts. 
Visually, the generated IR-drop map tracks the primary magnitude and spatial patterns of the ground truth, including prominent high-drop regions and surrounding gradients.

\section{Conclusion}
We introduced GIF, a conditional diffusion framework for IR-drop map generation that uses geometrical features and topological information as joint conditioning signals. By combining geometric features, a netlist-level graph encoded into graph features using multimodal fusion on a diffusion U-Net. 
GIF captures both local IR-drop variation and long-range dependencies.
The framework achieves strong accuracy on CircuitNet-N28 and stable correlation on the extended CircuitNet-N14 setup, showing that diffusion models can effectively integrate image and graph-based conditioning for IR-drop analysis. This formulation provides a scalable alternative to traditional IR-drop workflows and opens opportunities for future work in multi-scale conditioning, dynamic power information, and cross-technology generalization. In addition to these gains, GIF establishes a unified generative formulation for IR-drop, which has not been explored in prior work. By enabling IR-drop maps to be synthesized rather than directly regressed, the framework offers a foundation for future benchmarking and for building richer frameworks around generative modeling.
\bibliographystyle{splncs04}
\bibliography{main}

@String(AAAI  = {AAAI})

@article{ho2020denoising,
  title={Denoising diffusion probabilistic models},
  author={Ho, Jonathan and Jain, Ajay and Abbeel, Pieter},
  journal={Advances in neural information processing systems},
  volume={33},
  pages={6840--6851},
  year={2020}
}

@article{dosovitskiy2020image,
  title={An image is worth 16x16 words: Transformers for image recognition at scale},
  author={Dosovitskiy, Alexey},
  journal={arXiv preprint arXiv:2010.11929},
  year={2020}
}

@article{kingma2013auto,
  title={Auto-encoding variational bayes},
  author={Kingma, Diederik P and Welling, Max},
  journal={arXiv preprint arXiv:1312.6114},
  year={2013}
}

@inproceedings{rombach2022high,
  title={High-resolution image synthesis with latent diffusion models},
  author={Rombach, Robin and Blattmann, Andreas and Lorenz, Dominik and Esser, Patrick and Ommer, Bj{\"o}rn},
  booktitle={Proceedings of the IEEE/CVF conference on computer vision and pattern recognition},
  pages={10684--10695},
  year={2022}
}

@inproceedings{liu2021swin,
  title={Swin transformer: Hierarchical vision transformer using shifted windows},
  author={Liu, Ze and Lin, Yutong and Cao, Yue and Hu, Han and Wei, Yixuan and Zhang, Zheng and Lin, Stephen and Guo, Baining},
  booktitle={Proceedings of the IEEE/CVF international conference on computer vision},
  pages={10012--10022},
  year={2021}
}

@inproceedings{zheng2023lay,
  title={Lay-net: Grafting netlist knowledge on layout-based congestion prediction},
  author={Zheng, Su and Zou, Lancheng and Xu, Peng and Liu, Siting and Yu, Bei and Wong, Martin},
  booktitle={2023 IEEE/ACM International Conference on Computer Aided Design (ICCAD)},
  pages={1--9},
  year={2023},
  organization={IEEE}
}

@inproceedings{fang2018machine,
  title={Machine-learning-based dynamic IR drop prediction for ECO},
  author={Fang, Yen-Chun and Lin, Heng-Yi and Sui, Min-Yan and Li, Chien-Mo and Fang, Eric Jia-Wei},
  booktitle={2018 IEEE/ACM International Conference on Computer-Aided Design (ICCAD)},
  pages={1--7},
  year={2018},
  organization={IEEE}
}

@inproceedings{ho2019incpird,
  title={IncPIRD: Fast learning-based prediction of incremental IR drop},
  author={Ho, Chia-Tung and Kahng, Andrew B},
  booktitle={2019 IEEE/ACM International Conference on Computer-Aided Design (ICCAD)},
  pages={1--8},
  year={2019},
  organization={IEEE}
}

@inproceedings{xie2020fast,
  title={Fast IR Drop Estimation with Machine Learning},
  author={Xie, Zhiyao and Li, Hai and Xu, Xiaoqing and Hu, Jiang and Chen, Yiran},
  booktitle={2020 IEEE/ACM International Conference on Computer-Aided Design (ICCAD)},
  pages={1--8},
  year={2020},
  organization={IEEE}
}

@misc{zhao2024pdnnetpdnawaregnncnnheterogeneous,
      title={PDNNet: PDN-Aware GNN-CNN Heterogeneous Network for Dynamic IR Drop Prediction}, 
      author={Yuxiang Zhao and Zhuomin Chai and Xun Jiang and Yibo Lin and Runsheng Wang and Ru Huang},
      year={2024},
      eprint={2403.18569},
      archivePrefix={arXiv},
      primaryClass={cs.LG},
      url={https://arxiv.org/abs/2403.18569}, 
}

@inproceedings{chhabria2021mavirec,
  title={MAVIREC: ML-aided vectored IR-drop estimation and classification},
  author={Chhabria, Vidya A and Zhang, Yanqing and Ren, Haoxing and Keller, Ben and Khailany, Brucek and Sapatnekar, Sachin S},
  booktitle={2021 Design, Automation \& Test in Europe Conference \& Exhibition (DATE)},
  pages={1825--1828},
  year={2021},
  organization={IEEE}
}

@INPROCEEDINGS{9045574,
  author={Xie, Zhiyao and Ren, Haoxing and Khailany, Brucek and Sheng, Ye and Santosh, Santosh and Hu, Jiang and Chen, Yiran},
  booktitle={2020 25th Asia and South Pacific Design Automation Conference (ASP-DAC)}, 
  title={PowerNet: Transferable Dynamic IR Drop Estimation via Maximum Convolutional Neural Network}, 
  year={2020},
  volume={},
  number={},
  pages={13-18},
  doi={10.1109/ASP-DAC47756.2020.9045574}}

@INPROCEEDINGS{9045303,
  author={Chhabria, Vidya A. and Kahng, Andrew B. and Kim, Minsoo and Mallappa, Uday and Sapatnekar, Sachin S. and Xu, Bangqi},
  booktitle={2020 25th Asia and South Pacific Design Automation Conference (ASP-DAC)}, 
  title={Template-based PDN Synthesis in Floorplan and Placement Using Classifier and CNN Techniques}, 
  year={2020},
  volume={},
  number={},
  pages={44-49},
  doi={10.1109/ASP-DAC47756.2020.9045303}}

@article{thoratgroot,
  title={GROOT: Graph Edge Re-growth and Partitioning for the Verification of Large Designs in Logic Synthesis},
  author={Thorat, Kiran and Peng, Hongwu and Luo, Yuebo and Xie, Xi and Huang, Shaoyi and Hasan, Amit and Zhao, Jiahui and Li, Yingjie and Wu, Nan and Shi, Zhijie and others}
}

@inproceedings{Xie_2020,
   title={PowerNet: Transferable Dynamic IR Drop Estimation via Maximum Convolutional Neural Network},
   url={http://dx.doi.org/10.1109/ASP-DAC47756.2020.9045574},
   DOI={10.1109/asp-dac47756.2020.9045574},
   booktitle={2020 25th Asia and South Pacific Design Automation Conference (ASP-DAC)},
   publisher={IEEE},
   author={Xie, Zhiyao and Ren, Haoxing and Khailany, Brucek and Sheng, Ye and Santosh, Santosh and Hu, Jiang and Chen, Yiran},
   year={2020},
   month=jan, pages={13–18} }

@inproceedings{zhong2005fast,
  title={Fast algorithms for IR drop analysis in large power grid},
  author={Zhong, Yu and Wong, Martin DF},
  booktitle={ICCAD-2005. IEEE/ACM International Conference on Computer-Aided Design, 2005.},
  pages={351--357},
  year={2005},
  organization={IEEE}
}

@inproceedings{kose2011fast,
  title={Fast algorithms for IR voltage drop analysis exploiting locality},
  author={K{\"o}se, Sel{\c{c}}uk and Friedman, Eby G},
  booktitle={Proceedings of the 48th Design Automation Conference},
  pages={996--1001},
  year={2011}
}

@misc{ronneberger2015unetconvolutionalnetworksbiomedical,
      title={U-Net: Convolutional Networks for Biomedical Image Segmentation}, 
      author={Olaf Ronneberger and Philipp Fischer and Thomas Brox},
      year={2015},
      eprint={1505.04597},
      archivePrefix={arXiv},
      primaryClass={cs.CV},
      url={https://arxiv.org/abs/1505.04597}, 
}

@article{fardo2016formal,
  title={A formal evaluation of PSNR as quality measurement parameter for image segmentation algorithms},
  author={Fardo, Fernando A and Conforto, Victor H and De Oliveira, Francisco C and Rodrigues, Paulo S},
  journal={arXiv preprint arXiv:1605.07116},
  year={2016}
}

@ARTICLE{1284395,
  author={Zhou Wang and Bovik, A.C. and Sheikh, H.R. and Simoncelli, E.P.},
  journal={IEEE Transactions on Image Processing}, 
  title={Image quality assessment: from error visibility to structural similarity}, 
  year={2004},
  volume={13},
  number={4},
  pages={600-612},
  doi={10.1109/TIP.2003.819861}}

@article{chai2023circuitnet,
  title={Circuitnet: An open-source dataset for machine learning in vlsi cad applications with improved domain-specific evaluation metric and learning strategies},
  author={Chai, Zhuomin and Zhao, Yuxiang and Liu, Wei and Lin, Yibo and Wang, Runsheng and Huang, Ru},
  journal={IEEE Transactions on Computer-Aided Design of Integrated Circuits and Systems},
  volume={42},
  number={12},
  pages={5034--5047},
  year={2023},
  publisher={IEEE}
}

@inproceedings{jiang2024circuitnet,
  title={Circuitnet 2.0: An advanced dataset for promoting machine learning innovations in realistic chip design environment},
  author={Jiang, Xun and Chai, Zhuomin and Zhao, Yuxiang and Lin, Yibo and Wang, Runsheng and Huang, Ru and others},
  booktitle={The Twelfth International Conference on Learning Representations},
  year={2024}
}

@article{yang2022versatile,
  title={Versatile multi-stage graph neural network for circuit representation},
  author={Yang, Shuwen and Yang, Zhihao and Li, Dong and Zhang, Yingxueff and Zhang, Zhanguang and Song, Guojie and Hao, Jianye},
  journal={Advances in Neural Information Processing Systems},
  volume={35},
  pages={20313--20324},
  year={2022}
}

@inproceedings{perez2018film,
  title={Film: Visual reasoning with a general conditioning layer},
  author={Perez, Ethan and Strub, Florian and De Vries, Harm and Dumoulin, Vincent and Courville, Aaron},
  booktitle={Proceedings of the AAAI conference on artificial intelligence},
  volume={32},
  number={1},
  year={2018}
}

@inproceedings{huang2017arbitrary,
  title={Arbitrary style transfer in real-time with adaptive instance normalization},
  author={Huang, Xun and Belongie, Serge},
  booktitle={Proceedings of the IEEE international conference on computer vision},
  pages={1501--1510},
  year={2017}
}

@book{wolf2008modern,
  title={Modern VLSI Design: Systems on Silicon},
  author={Wolf, Wayne},
  year={2008},
  publisher={Pearson}
}

@book{sherwani1995algorithms,
  title={Algorithms for VLSI Physical Design Automation},
  author={Sherwani, Naveed},
  year={1995},
  publisher={Springer}
}

@inproceedings{nassif2001modeling,
  title={Modeling and analysis of manufacturing variations},
  author={Nassif, Sani R},
  booktitle={Proceedings of the IEEE 2001 Custom Integrated Circuits Conference (Cat. No. 01CH37169)},
  pages={223--228},
  year={2001},
  organization={IEEE}
}

@inproceedings{chen1997power,
  title={Power supply noise analysis methodology for deep-submicron VLSI chip design},
  author={Chen, Howard H and Ling, David D},
  booktitle={Proceedings of the 34th annual Design Automation Conference},
  pages={638--643},
  year={1997}
}

@article{borkar2002design,
  title={Design challenges of technology scaling},
  author={Borkar, Shekhar},
  journal={IEEE micro},
  volume={19},
  number={4},
  pages={23--29},
  year={2002},
  publisher={IEEE}
}

@inproceedings{fatima2023analysis,
  title={Analysis of ir drop for robust power grid of semiconductor chip design: a review},
  author={Fatima, Bushra and Chandel, Rajeevan},
  booktitle={ITM web of conferences},
  volume={54},
  pages={04001},
  year={2023},
  organization={EDP Sciences}
}
\clearpage
\setcounter{page}{1}
\title{Supplementary Material}
\maketitlesupplementary
\vspace{-5mm}
\begin{center}
{\large GIF: A Conditional Multimodal Generative Framework for IR Drop Imaging in Chip Layouts}
\end{center}

\appendix
\setcounter{section}{0}
\setcounter{subsection}{0}
\setcounter{figure}{0}
\setcounter{table}{0}
\setcounter{equation}{0}

\renewcommand{\thesection}{\Alph{section}}
\renewcommand{\thesubsection}{\thesection.\arabic{subsection}}
\renewcommand{\thefigure}{\thesection.\arabic{figure}}
\renewcommand{\thetable}{\thesection.\arabic{table}}
\renewcommand{\theequation}{\thesection.\arabic{equation}}

% important for hyperref:
\renewcommand{\theHsection}{supp.\Alph{section}}
\renewcommand{\theHsubsection}{supp.\Alph{section}.\arabic{subsection}}
\renewcommand{\theHfigure}{supp.\Alph{section}.\arabic{figure}}
\renewcommand{\theHtable}{supp.\Alph{section}.\arabic{table}}
\renewcommand{\theHequation}{supp.\Alph{section}.\arabic{equation}}

\section{Background: Modern Chip Design Flow and IR-Drop}
\label{sec:supp_background}
\begin{figure}[t]
    \centering
    \includegraphics[width=0.6\linewidth]{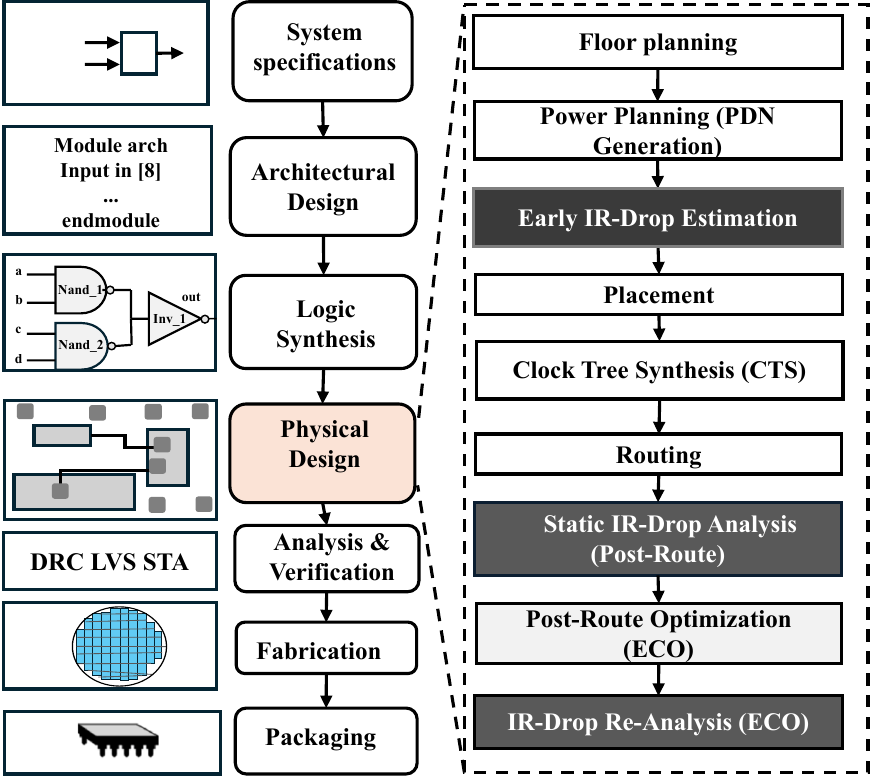}
    \caption{Modern chip design flow including physical design (orange) 
    and IR-drop analysis (black).}
    \label{fig:eda_flow}
\end{figure}

Figure~\ref{fig:eda_flow} shows modern chip design follows a standard sequence of stages including system specification, architecture, RTL, logic synthesis, physical design, and sign-off. Early stages establish the architecture and generate the logic, while physical design resolves the geometric and electrical constraints that govern on-chip behavior~\cite{wolf2008modern,sherwani1995algorithms}. This stage determines where cells are placed, how wires are routed, and how the power-delivery network (PDN) distributes supply voltage across increasingly resistive metal layers. As technology scales, wire resistance rises sharply and switching demand grows, making stable power delivery a central limitation in advanced designs~\cite{borkar2002design, fatima2023analysis}. As illustrated in Figure~\ref{fig:eda_flow}, these stages collectively shape the final layout and its power-distribution characteristics.

IR-drop is the reduction in supply voltage caused by current flowing through the resistive PDN. Although conceptually simple, its consequences are severe: small voltage reductions produce measurable delay shifts, alter noise margins, and destabilize logic under high activity~\cite{nassif2001modeling,chen1997power}. IR-drop interacts with clock uncertainty and process variation, amplifying timing sensitivity in deeply scaled nodes. As a result, IR-drop has become a dominant factor in design closure, often determining whether a layout can meet its timing targets after routing~\cite{xie2020fast}. When violations are detected late typically during Static IR-Drop Analysis in the post-route stage they trigger engineering change order (ECO) loops that require re-routing, PDN reinforcement, or placement adjustments, all of which propagate through timing and congestion, significantly increasing turnaround time.

Reliable IR-drop generation is therefore essential to reduce costly iterations and guide early design choices. Fast, accurate generation of IR-drop maps enables designers to identify weak PDN regions, anticipate voltage-loss patterns, and evaluate design modifications without repeatedly invoking expensive sign-off tools. By exposing critical hotspots earlier in the flow, generative models help stabilize timing behavior and improve the likelihood that the final layout converges without disruptive late-stage modifications.

\section{Image Graph Fusion Mechanism}

\begin{figure}[h]
    \centering
    \includegraphics[width=0.85\linewidth]{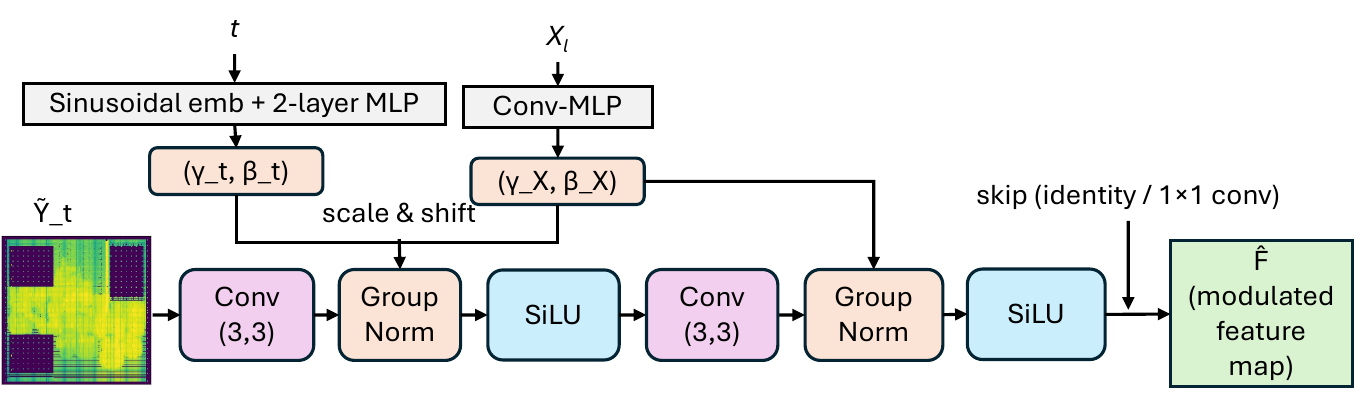}
    \caption{ResBlock with FiLM conditioning. Timestep embedding and
    geometric layout features $X_\ell$ produce affine scale shift pairs
    applied sequentially after each GroupNorm.}
    \label{fig:supp_resblock}
\end{figure}

Figure~\ref{fig:supp_resblock} shows the internal structure of the ResBlock
applied at every level of the denoising UNet. The input feature map
$F^{(\ell)}$ passes through two $3{\times}3$ convolution, GroupNorm, SiLU
in sequences. The timestep conditioning $(\gamma_t, \beta_t)$, produced by a
sinusoidal embedding followed by a two-layer MLP, is applied as a scale and
shift after the first GroupNorm. The geometry conditioning
$(\gamma_X^{(\ell)}, \beta_X^{(\ell)})$, produced by a convolutional MLP
applied to the downsampled layout features $X_\ell$, is applied after the
second GroupNorm. A skip connection (identity or $1{\times}1$ convolution)
is added to the output:
\[
\widehat{F}^{(\ell)}
    = \gamma^{(\ell)}(X_\ell, t) \odot F^{(\ell)}
      + \beta^{(\ell)}(X_\ell, t).
\]

\begin{figure}[h]
    \centering
    \includegraphics[width=0.8\linewidth]{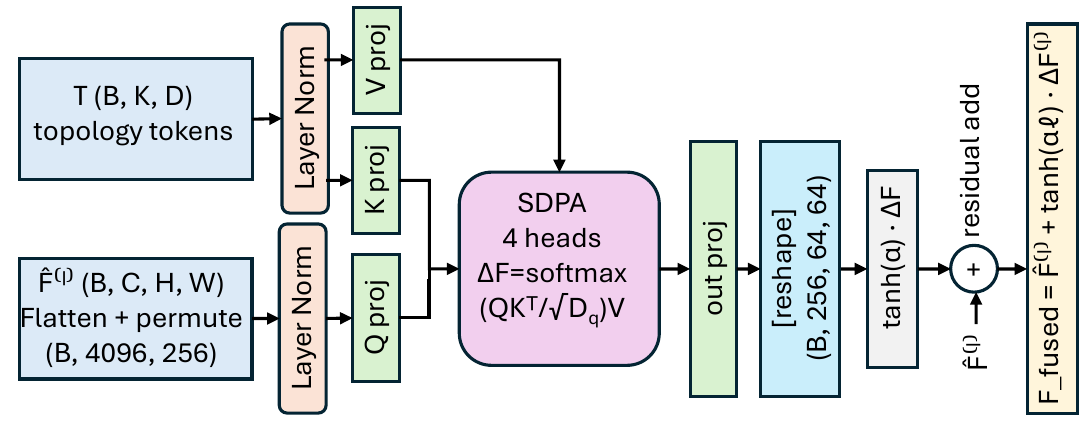}
    \caption{Cross-attention injection at the bottleneck. Topology tokens
    $T$ serve as keys and values while spatial features serve as queries.
    The gate $\tanh(\alpha_\ell)$, initialized to zero, controls the
    contribution of the topology signal.}
    \label{fig:supp_crossattn}
\end{figure}

Figure~\ref{fig:supp_crossattn} shows the cross-attention injection at the
bottleneck ($64{\times}64$, 256 channels). The feature map
$\widehat{F}^{(\ell)}$ is flattened to $(B,\, H_\ell W_\ell,\, C)$, passed
through LayerNorm, and projected to queries $Q^{(\ell)}$. The topology
tokens $T \in \mathbb{R}^{K \times D}$ are passed through a separate
LayerNorm and projected to keys $K$ and values $V$. Scaled dot-product
attention with 4 heads produces:
\[
\Delta F^{(\ell)}
=
\mathrm{softmax}\!\left(
    \frac{Q^{(\ell)} K^\top}{\sqrt{D_q}}
  \right) V.
\]
The output is projected and reshaped to $(B,\, C,\, H_\ell,\, W_\ell)$,
then added via a scalar gate $\alpha_\ell$ initialized to zero:
\[
F_{\text{fused}}^{(\ell)}
    = \widehat{F}^{(\ell)}
    + \tanh(\alpha_\ell)\,
      \mathrm{reshape}\!\left(\Delta F^{(\ell)}\right).
\]
\section{Dataset Details} 

% \textcolor{red}{@mustfa please read double check an the dataset details and decide whether we should keep the table}

\noindent \textbf{CircuitNet-N28}
CircuitNet-N28 provides tile-based physical design features and IR-drop maps generated from six RTL designs. Each layout is discretized into a regular grid of approximately $300\times 300$ tiles, where each tile corresponds to a $2.25\,\mu\mathrm{m}\times 2.25\,\mu\mathrm{m}$ region of the physical layout (chip area $\sim 450\,\mu\mathrm{m}\times 450\,\mu\mathrm{m}$). This yields spatial feature maps that directly encode routing demand, placement density, switching activity, and power-related signals.  
We use all 10{,}242 samples from CircuitNet-N28 and follow the official design-wise split: four designs for training, one for validation, and one for testing.
% (see Table~\ref{tab:dataset_split}).
All feature and IR-drop maps are resized to $256\times256$ and normalized before training.

% \begin{table}[t]
% \centering
% \caption{Design-wise split of the CircuitNet-N28 dataset.}
% \label{tab:dataset_split}
% \small
% \begin{tabular}{lcc}
% \toprule
% \textbf{Design} & \textbf{Split} & \textbf{Samples} \\
% \midrule
% RISCY-a        & Train & 2{,}003 \\
% RISCY-b        & Train & 1{,}858 \\
% RISCY-FPU-a    & Train & 1{,}969 \\
% zero-riscy-a   & Train & 2{,}042 \\
% \midrule
% RISCY-FPU-b    & Val   & 1{,}248 \\
% zero-riscy-b   & Test  & 1{,}122 \\
% \midrule
% \textbf{Total} &       & \textbf{10{,}242} \\
% \bottomrule
% \end{tabular}
% \end{table}

\noindent \textbf{CircuitNet-N14}
CircuitNet-N14 contains full-chip IR-drop maps and physical design features extracted from eight RTL designs. Compared to N28, the N14 layouts exhibit larger variation in floorplan styles, utilization, aspect ratio, and power delivery configurations. Each layout is represented by a spatial grid of comparable resolution (on the order of $300\times 300$ tiles), covering the entire chip image at a uniform sampling density.  
We use all 10{,}444 available samples and adopt the design-wise split provided by CircuitNet: six designs for training, Vortex-small for validation, and zero-riscy for testing.

\section{Evaluation Metrics}
IR-drop analysis in physical design has one practical goal: identify where
voltage drops are large enough to cause timing failures or functional
errors, so that the power delivery network can be strengthened before
tapeout. The metrics we report are chosen to reflect this goal directly.
Let $\mathbf{Y} \in \mathbb{R}^{H \times W}$ denote the ground-truth
IR-drop map and $\widehat{\mathbf{Y}} \in \mathbb{R}^{H \times W}$ the
generated map, both normalized to $[0,1]$ relative to the supply voltage.
We flatten both into vectors $y, \hat{y} \in \mathbb{R}^{N}$, $N = HW$.

\noindent \textbf{Pixel-level accuracy.}
MAE measures the average per-pixel voltage error across the layout:
\begin{equation}
\mathrm{MAE} = \frac{1}{N} \sum_{i=1}^{N} \bigl| \hat{y}_{i} - y_{i} \bigr|.
\end{equation}
Because IR-drop values are normalized to $[0,1]$ relative to the supply
voltage, MAE is directly interpretable: an MAE of 0.026 means the model
is on average 2.6\% of the supply voltage away from the ground truth at
each layout location. RMSE places higher weight on large errors:
\begin{equation}
\mathrm{RMSE} = \sqrt{ \frac{1}{N} \sum_{i=1}^{N} \left( \hat{y}_{i} -
y_{i} \right)^{2} }.
\end{equation}
In IR-drop analysis, large errors matter disproportionately because an
underestimated hotspot that crosses the timing margin threshold is a
functional failure. RMSE is therefore a more safety-relevant measure than
MAE alone. PSNR is included for consistency with the CircuitNet benchmark
protocol~\cite{zhao2024pdnnetpdnawaregnncnnheterogeneous,
chhabria2021mavirec} and does not carry additional physical interpretation
beyond its relationship to MSE.

\begin{figure}[h]
    \centering
    \includegraphics[width=0.95\linewidth]{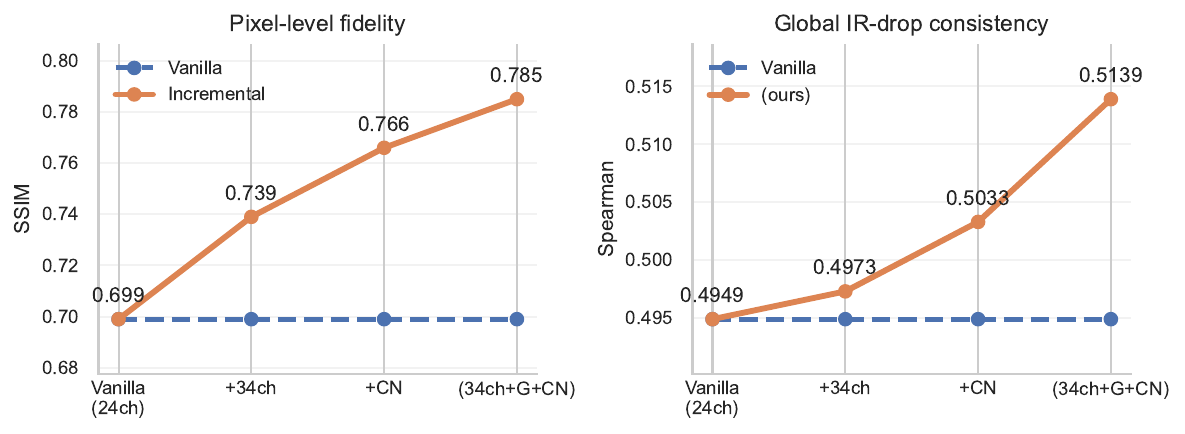}
    \caption{SSIM and Spearman correlation across incremental model 
    configurations on CircuitNet-N28. \textit{Left}: SSIM improves as 
    each component is added, reflecting improved spatial structure of the 
    generated IR-drop map. \textit{Right}: Spearman correlation across 
    the same steps, showing improved hotspot severity ordering with each 
    added component.}
    \label{fig:supp_ablation}
\end{figure}

\noindent \textbf{Spatial structure fidelity.}
SSIM evaluates local luminance, contrast, and structural agreement across
windowed regions of the layout~\cite{1284395}:
\begin{equation}
\mathrm{SSIM}(\mathbf{Y}, \widehat{\mathbf{Y}})
= \frac{
(2\mu_{y}\mu_{\hat{y}} + C_{1})(2\sigma_{y\hat{y}} + C_{2})
}{
(\mu_{y}^{2} + \mu_{\hat{y}}^{2} + C_{1})
(\sigma_{y}^{2}  + \sigma_{\hat{y}}^{2}  + C_{2})
},
\end{equation}
computed with an $11{\times}11$ Gaussian window~\cite{fardo2016formal}.
IR-drop maps are not arbitrary images. The spatial gradients in the
voltage map reflect how current flows through the resistive power mesh
from supply rails to standard cells. A generated map that places hotspots
in the right locations but produces abrupt or noisy voltage transitions
would be physically incorrect, because real current flow through a
resistive network produces smooth, continuous gradients. SSIM captures
exactly this: a model that smears or sharpens voltage gradients
incorrectly will score low on SSIM even if per-pixel errors are small.
This makes SSIM a meaningful proxy for physical plausibility of the
generated IR-drop map, beyond what MAE and RMSE can capture.
Figure~\ref{fig:supp_ablation} (left) shows SSIM improving monotonically
as each component of GIF is added.

\noindent \textbf{Hotspot severity consistency.}
Pearson correlation measures whether the generated map reproduces the
global IR-drop profile of the layout, that is, whether regions of
elevated voltage drop in the ground truth are also predicted as high by
the model:
\begin{equation}
\mathrm{Pearson}
= \frac{
\sum_{i=1}^{N} (\hat{y}_{i} - \bar{\hat{y}})(y_{i} - \bar{y})
}{
\sqrt{\sum_{i=1}^{N}(\hat{y}_{i} - \bar{\hat{y}})^{2}}\,
\sqrt{\sum_{i=1}^{N}(y_{i} - \bar{y})^{2}}
}.
\end{equation}
Spearman correlation measures whether the model correctly orders layout
regions from least to most critical, independent of absolute voltage
values:
\begin{equation}
\mathrm{Spearman} = \mathrm{Pearson}\!\left( \mathrm{rank}(\hat{y}),\,
\mathrm{rank}(y) \right).
\end{equation}
This distinction matters in practice. A PDN engineer does not inspect
every pixel of an IR-drop map. The standard workflow is to identify the
worst hotspot regions, ranked by severity, and decide where to add power
stripes or decoupling capacitors. A model with high Pearson but low
Spearman reproduces the overall voltage distribution but fails to rank
individual hotspot regions correctly, which is a practically important
failure mode. Reporting both metrics together gives a complete picture of
whether the generated map is reliable for this decision process. Neither
Pearson nor Spearman is reported by prior IR-drop prediction
work~\cite{zhao2024pdnnetpdnawaregnncnnheterogeneous,
chhabria2021mavirec}; we include both because they directly reflect the
engineering use case that motivates IR-drop analysis.
Figure~\ref{fig:supp_ablation} (right) shows Spearman across ablation
steps.

\noindent \textbf{Evaluation protocol.}
All metrics are computed on a single generated sample per test layout.
In IR-drop analysis, the ground truth is produced by a single
deterministic simulation from a commercial EDA tool given a fixed layout
and power map. There is no distribution of ground truths. The task is to
predict what that simulation would produce, and a single generated sample
is the appropriate unit of comparison. This is the same protocol used by
all prior discriminative
methods~\cite{zhao2024pdnnetpdnawaregnncnnheterogeneous,
chhabria2021mavirec}, and it is the correct protocol for this task
regardless of whether the model is generative or deterministic. GIF
requires 1000 denoising steps per sample, so generating multiple samples
per layout for the full test set is also computationally prohibitive.

\section{Additional Analysis on CircuitNet-N14 Generation}
\label{sec:supp_additional_analysis}
\noindent \textbf{First IR-Drop Results on CircuitNet-N14.}
To the best of our knowledge, GIF provides the first reported IR-drop results of any kind on CircuitNet-N14.  
Although the dataset includes IR-drop ground truth, no prior prediction (CNN, GNN, or hybrid), or generative  (diffusion-based) method has published evaluation results on CircuitNet-N14 dataset.  
Thus, the values presented in the main paper represent the initial quantitative baselines for IR-drop map generation at 14\,nm.  
These results demonstrate that the diffusion backbone maintains strong correlation behavior even under the much higher variability of N14 physical layouts.

\vspace{0.5em}
\noindent \textbf{Graph Availability in CircuitNet-N14.}
As described in the supplementary implementation details, graphs for N14 are constructed using the same procedure as for N28.  
However, CircuitNet-N14 contains a large number of placement snapshots whose instance bounding boxes are all zero.  
These samples cannot produce meaningful geometric node features and must be skipped during graph construction.  
This behavior arises from irregularities in the dataset metadata and is not specific to our framework.

Table~\ref{tab:n14_graph_stats} reports the number of valid graphs extracted for each design family.  
Several families (e.g., \texttt{Vortex-small}) yield no usable graphs, while others produce only a limited number.  
Since the full N14 feature/label set contains 10{,}444 IR-drop samples, many conditioning instances necessarily receive no graph tokens.  
This imbalance explains why graph-conditioned variants offer limited benefit or slight degradation on N14, in contrast to N28 where graph metadata are complete and consistently available.

\begin{table}[h]
\centering
\caption{Number of valid graphs constructed for CircuitNet-N14 (from the official metadata).  
Designs with all-zero placements produce no usable graphs.}
\label{tab:n14_graph_stats}
\small
\begin{tabular}{lcc}
\toprule
\textbf{Design Family} & \textbf{Valid Graphs} & \textbf{Total Placements} \\
\midrule
RISCY            & 3373 & 3456 \\
RISCY-FPU        & 3456 & 3456 \\
Vortex-large     & 62   & 74   \\
Vortex-small     & 0    & 96   \\
nvdia-large      & 54   & 68   \\
nvdia-small      & 88   & 93   \\
openc910-1       & 96   & 96   \\
zero-riscy       & 3456 & 3456 \\
\midrule
\textbf{Total}   & 10585 & 10795 \\
\bottomrule
\end{tabular}
\end{table}

% \noindent \textbf{Summary.}
GIF establishes the first IR-drop benchmarks on CircuitNet-N14 and maintains strong correlation consistency despite the dataset’s higher layout variability.  
The reduced benefit of graph conditioning on N14 is driven by incomplete and inconsistent metadata, not by limitations of the fusion design.  
Where graph features are reliably available (e.g., N28), graph conditioning consistently improves performance.

\end{document}